\documentclass[conference]{IEEEtran}

\usepackage{xcolor}
\usepackage[pdftex]{graphicx}
\usepackage{subfigure}
\usepackage{makecell}
\usepackage{amssymb}
\usepackage{amsmath}
\usepackage{bm}
\usepackage[pagebackref,breaklinks,colorlinks,citecolor=blue]{hyperref}
\usepackage[linesnumbered,ruled,vlined]{algorithm2e}
\usepackage{subfiles}
\usepackage{siunitx}
\usepackage{multirow}

\title{FusionForce: End-to-end Differentiable Neural-Symbolic Layer for Trajectory Prediction}

\author{
  \IEEEauthorblockN{Ruslan Agishev, Karel Zimmermann}
  \IEEEauthorblockA{
    Faculty of Electrical Engineering, Czech Technical University in Prague\\
    Email: agishrus@fel.cvut.cz
  }
}

\begin{document}

\maketitle

\begin{abstract}
We propose end-to-end differentiable model that predicts robot trajectories on rough offroad terrain from camera images and/or lidar point clouds. %The proposed model enforces the laws of classical mechanics through a physics-aware neural symbolic layer while preserving the ability to learn from large-scale data.
The model integrates a learnable component that predicts robot-terrain interaction forces with a neural-symbolic layer that enforces the laws of classical mechanics and consequently improves generalization on out-of-distribution data.
%The learnable component is a multi-sensor fusion encoder that unifies point cloud and RGB features in bird-eye-view space.
The neural-symbolic layer includes a differentiable physics engine that computes the robot’s trajectory by querying these forces at
the points of contact with the terrain.
As the proposed architecture comprises substantial geometrical and physics priors,
the resulting model can also be seen as a learnable physics engine conditioned on real sensor data that delivers $10^4$ trajectories per second.
We argue and empirically demonstrate that this architecture reduces the sim-to-real gap and mitigates out-of-distribution sensitivity.
The differentiability, in conjunction with the rapid simulation speed, makes the model well-suited for various applications
including model predictive control, trajectory shooting, supervised and reinforcement learning, or SLAM.
%The codes and data are publicly available.
\end{abstract}


\section{Introduction}
\begin{figure}[ht!]
    \centering
    \includegraphics[width=0.45\textwidth]{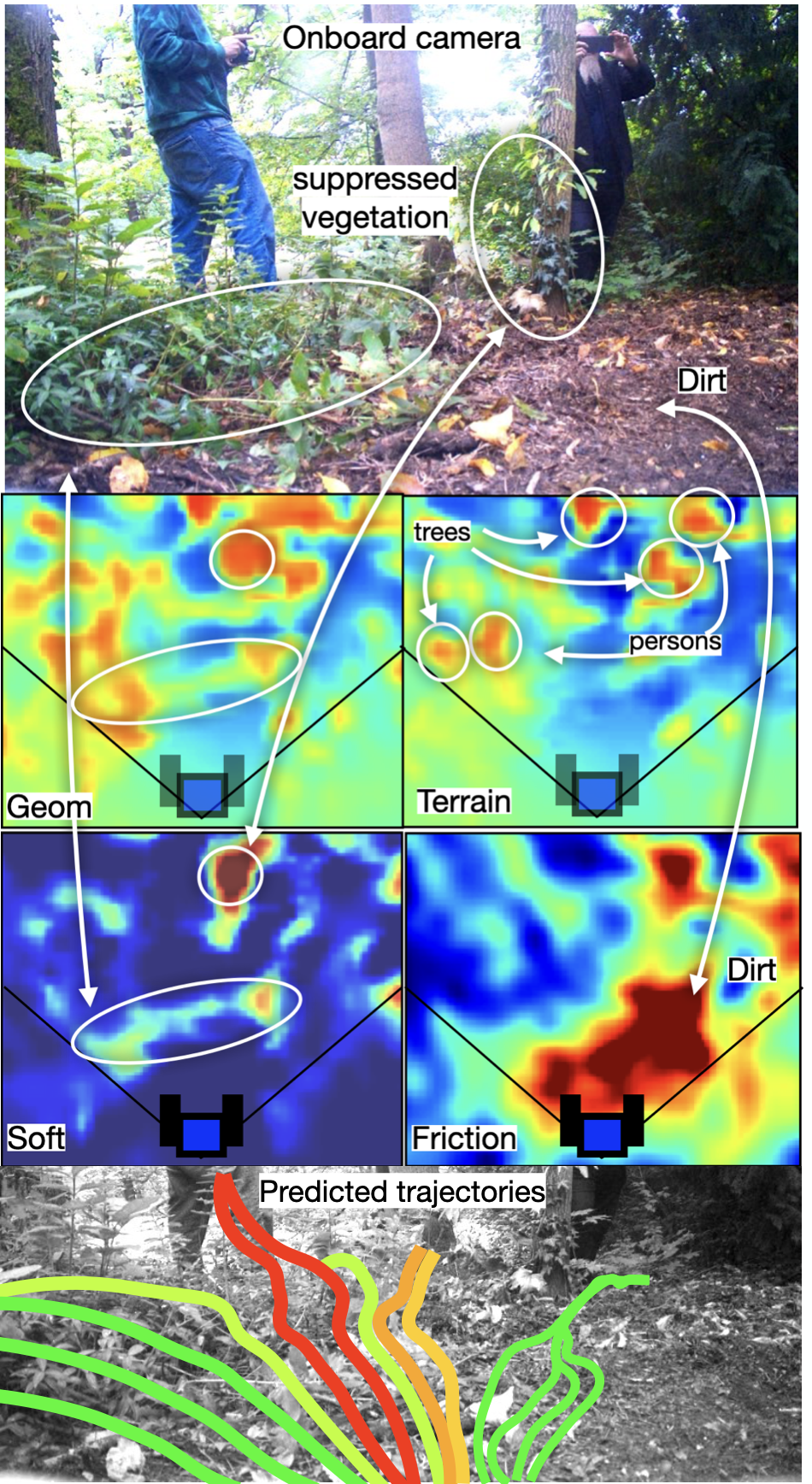}
    \caption{Camera, predicted terrain properties and trajectories.}
    %\includegraphics[width=0.44\textwidth]{imgs/monoforce_heightmaps}
    % \caption{\textbf{Predicted terrain properties:} Geometrical heightmap replicates lidar heightmap. Soft heightmap outlines soft parts of the terrain, such as vegetation. Supporting terrain map that depicts the terrain that generates robot-terrain interaction forces. Friction heightmap that allows to estimate track forces see significantly higher values on dirt.
    % }
    \label{fig:monoforce_heightmaps}
%\vspace{-1cm}
\end{figure}

Autonomous robotics in off-road environments holds immense promise for various applications,
including outdoor logistics, inspections, and forestry operations.
Yet, unlike systems designed for controlled environments, such as factories, off-road autonomous systems still remain
prohibitively immature for dependable deployment.
The main challenge that prevents reliable off-road deployment lies in the inability to predict the behavior
of the robot on the terrain with a sufficient accuracy.
Especially, the prediction of robot trajectory in tall vegetation or muddy and rocky terrains from camera images and lidar scans remains an unresolved problem in outdoor robotics. We address this issue by proposing an end-to-end differentiable physics-aware model that simultaneously predicts various terrain properties and potential robot trajectories, see Fig.~\ref{fig:monoforce_heightmaps} for an example.

\begin{figure*}[t!]
  \centering
  \includegraphics[width=0.98\textwidth]{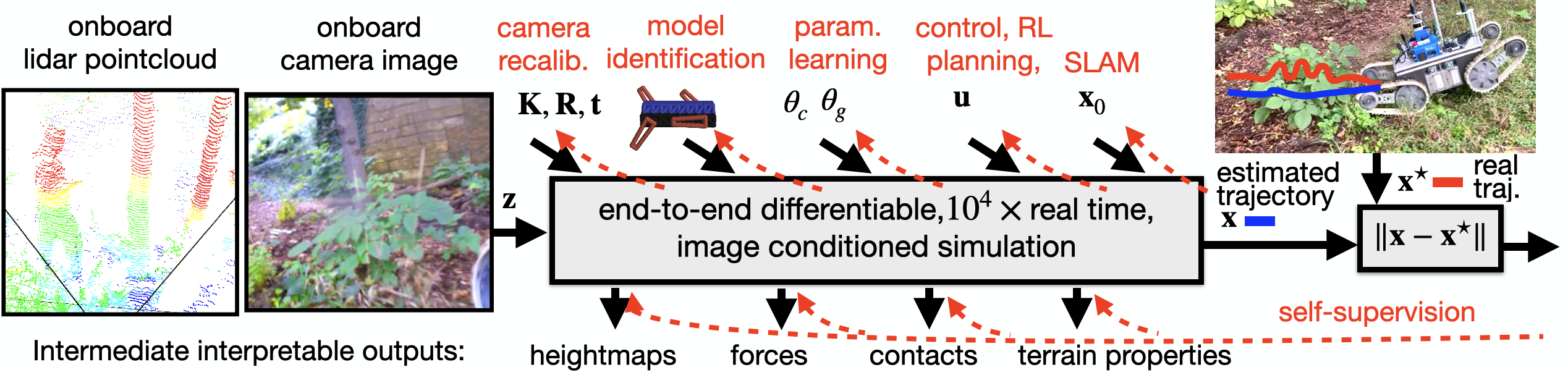}
  \caption{
  \textbf{Model overview:} The proposed model can be seen as an image-conditioned differentiable simulation that delivers a million simulated trajectories per second on the terrain depicted in the onboard camera image. The explainable structure also delivers many intermediate interpretable outputs that can serve for efficient self-supervision.}
  \label{fig:model_overview}  
\end{figure*}

\begin{figure}[t!]
  \centering
\includegraphics[width=0.46\textwidth]{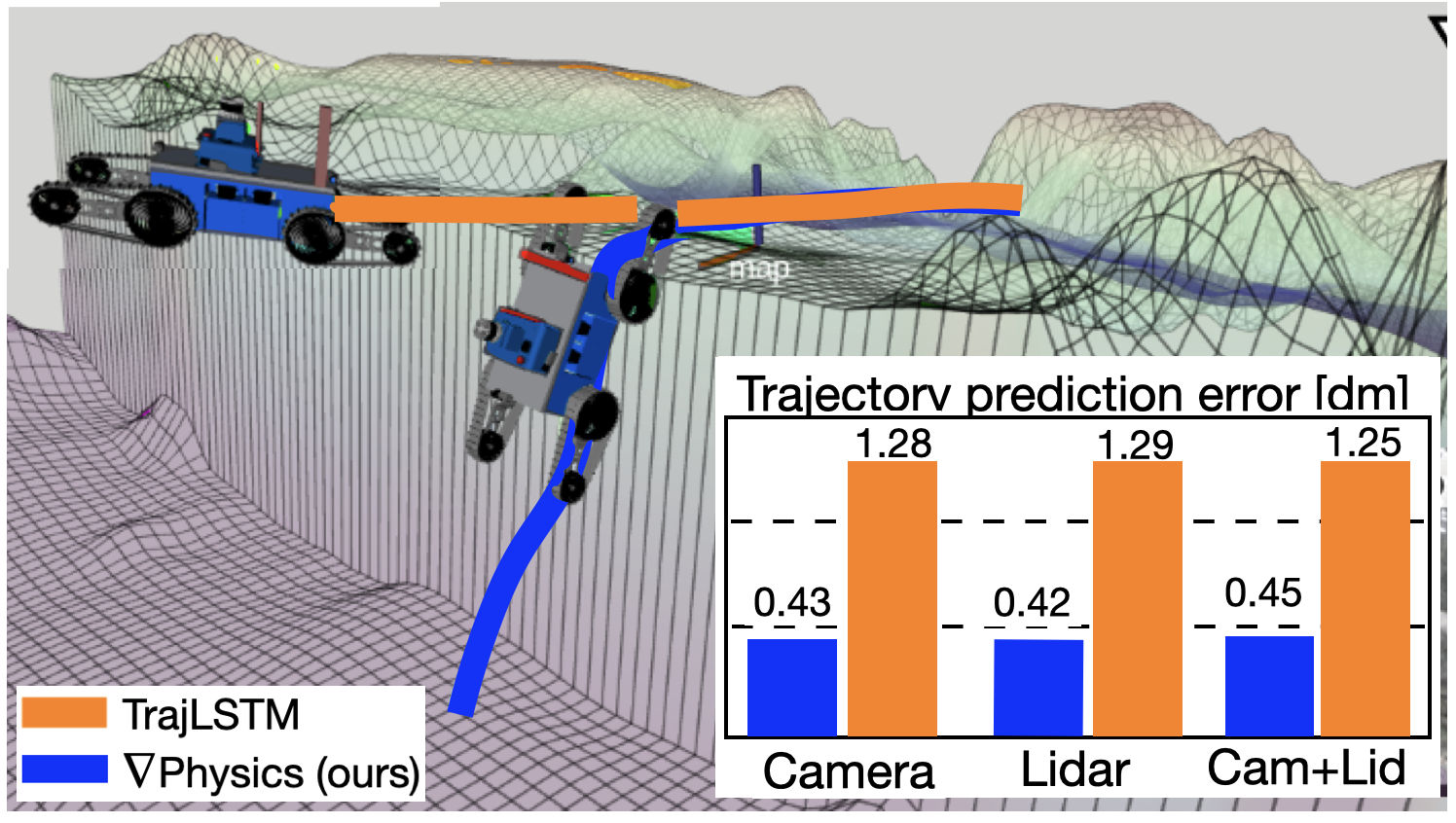}
  % \includegraphics[width=0.5\textwidth]{imgs/predictions/fusionforce_monoforce}
%  \caption{\textbf{Qualitative results:} The rows show (i) onboard camera images (the only input),
%    (ii) predicted supporting terrain and
%    (iii) a 3D view with predicted trajectory and visualized ground truth lidar point clouds
%    (not used for estimation).
%    While the majority of rigid objects are correctly reconstructed from the camera image,
%    robot flippers, tall grass, fallen branches, or soft hanging branches are correctly suppressed
%    as they do not influence the resulting robot trajectory.
%    Predicted terrain friction, dampening, and stiffness are not visualized for brevity.
%  }
  \caption{\textbf{Qualitative and quantitative results:} Comparison of the trajectory error with the state-of-the-art architecture based on the LSTM network. Proposed model achieves, due to the $\nabla$Physics layer, significantly better generalization on out-of-distribution data such as steep terrain cliffs.
  % to the previous MonoForce (RGB only input) version.
  % The rows show (i) onboard camera images,
  % (ii) FusionForce prediction: supporting terrain and robot's trajectory,
  % (iii) MonoForce prediction: supporting terrain and robot's trajectory (lidar cloud is given for reference),
  % The previous MonoForce model struggles to predict occluded trees and the distant scale is not output correctly
  % for some obstacles (the tree trunk in the bottom-right).
  % Predicted terrain friction, dampening, and stiffness are not visualized for brevity.
  }
  \label{fig:catch-eye}
\end{figure}

Over the last decade,
roboticists proposed a wide variety of \emph{white-box}~\cite{Fabian2020, Dogru-AuRo-2021, manoharan-IROS-2024}
and \emph{black-box} models~\cite{loquercio-ICRA-2023, Niu-FRAI-2023, Wellhausen-RAL-2019, hdif2023, kahn2020badgr} for the off-road trajectory prediction.
Black-box models primarily suffer from a severe out-of-distribution problem~-- the phenomenon
where the distribution of training data does not correspond to the testing data.
This problem naturally arises from the fact that the training data includes only the trajectories from safe
terrain traversals, such as small terrain steps, while the decision about terrain traversability also naturally
comprises the non-traversable terrains, such as tall cliffs, Fig.~\ref{fig:catch-eye}-top.
Since the step-to-cliff-generalization of large state-of-the-art black-box models is typically poor,
the safety of the resulting system is typically inferior.
% Unfortunately,
% the safety of the resulting system is mainly determined by the performance of the model on non-traversable terrains,
% which could be arbitrarily bad due to their absence in the training data.
On the other hand, \emph{white-box} architectures offer good generalization due to substantial inductive bias
but often suffer from the sim-to-real gap~-- the phenomenon where a model \("\)trained in\("\) or \("\)replaced by\("\)
simulation faces challenges or discrepancies when applied in the real world.
Although several techniques, such as rapid motor adaptation~\cite{rma-2021}, are known to partially suppress this issue
through domain randomization, the sim-to-real gap, as well as the out-of-distribution problem,
makes trajectory prediction from camera images prohibitively unreliable.
We introduce a \emph{grey-box} model that combines the best of both worlds to achieve better generalization and consequently better accuracy and smaller sim-to-real gap.
This model enforces white-box laws of classical mechanics through a physics-aware neural symbolic layer,
while preserving the ability to learn from huge data as it is end-to-end differentiable.
We demonstrate that independently of the employed input sensor (camera, lidar or both), the proposed neural-symbolic physics layer consistently improves trajectory prediction error when compared to a state-of-the-art data-driven architecture such as LSTM, see Fig.~\ref{fig:catch-eye} for an example. 
%even just a single onboard camera is enough for reliable autonomous deployment in off-road environments; see Figures~\ref{fig:catch-eye2}-bottom for some qualitative results and Fig.~\ref{fig:traj_errors} for quantitative evaluation.

%explainable, physics-aware, and end-to-end differentiable model that improves generalization and suppresses the sim-to-real gap to the level at which the camera input becomes more reliable than existing competitors.

%We introduce a \emph{grey-box}, explainable, physics-aware, and end-to-end differentiable model that improves generalization and suppresses the sim-to-real gap to the level at which the camera input becomes more reliable than existing competitors.
%(i) improves generalization by introducing inductive bias through neural symbolic layers and (ii) suppresses the sim-to-real gap through end-to-end learnability.

%substantial physics and geometrical priors encoded in the neural symbolic layers and the efficient self-supervision enabled by model explainability.

% included in the neural symbolic physics layer 
%enables self-supervised learning also from non-traversable terrains.

%%% STATISTICALLY INCONSISTENT LEARNING %%%%

%%% STRONG PRIORS, DIFFERENTIABILITY, APPLICATIONS, WARP/BRAX unstable gradients %%%
In our implementation, the black-box part of our model predicts robot-terrain interaction forces and the true height at which they appear,
while the neural symbolic layer, which contains a differentiable physics engine, solves the robot's trajectory
by querying these forces at the robot-terrain contacts.
As the proposed architecture comprises substantial geometrical and physics priors,
the resulting model can be also seen as \emph{learnable physics engine conditioned by a real image}
that delivers \emph{$10^4$ trajectories per second}; see~Fig.~\ref{fig:model_overview} for details.

In addition to that, the model is end-to-end-differentiable; therefore, gradients can be backpropagated towards its
(i) convolutional filters, (ii) camera and robot parameters, (iii) control, (iv) initial position, and (v) terrain properties.
The differentiability, in conjunction with the rapid simulation speed, makes the model suitable for a myriad of tasks,
including model predictive control~\cite{Amos-NEURIPS-2018}, trajectory shooting~\cite{Zeng-CVPR-2019},
supervised and reinforcement learning~\cite{schulman2017proximal}, SLAM~\cite{factorgraph-2017},
online robot-model reidentification or camera recalibration~\cite{Moravec-CVWW-2018}.
The explainable structure of the proposed architecture also delivers a variety of intermediate outputs,
such as terrain shape and its physical properties, robot-terrain reaction forces or contact points,
which can all serve as efficient sources of self-supervision if measured during the training set creation or restricted
in PINN-like manner~\cite{Farea-AI-2024}.

In addition to that, we observed that the instability of gradient computation makes existing differentiable simulators such as NVIDIA's WARP~\cite{warp2022} and Google's BRAX~\cite{brax2021} prohibitively unreliable
and slow for both learning and inference.
% We originally intended to implement the neural symbolic layer in existing differentiable simulators such as NVIDIA's WARP~\cite{warp2022} and Google's BRAX~\cite{brax2021}, however, we observed that the instability of gradient computation makes them prohibitively unreliable
% and slow for both the learning and the inference.
%Despite the existence of differentiable simulators such as NVIDIA's WARP~\cite{warp2022} and Google's BRAX~\cite{brax2021},
%we observed
%that the instability of gradient computation in all available solvers makes them prohibitively unreliable and slow for both the learning and the inference.
As a suitable replacement, we propose our own differentiable neural symbolic layer, which integrates
from-scratch implemented physics engine, into the proposed architecture.
This solution outperforms existing works in gradient stability and computational speed
while preserving sufficient accuracy
for reliable trajectory prediction.
As all target applications, including learning, control, planning, and SLAM, can be naturally parallelized,
we also achieve significant speed-up through massive parallelization on GPU.

We emphasize that all state-of-the-art models for trajectory prediction of ground robots, including ours, are inherently statistically inconsistent because training data cannot, in principle, cover hazardous scenarios. As a result, the common practice of using a near-infinite number of training examples, as seen in large foundation models~\cite{Kirillov-ICCV-2023}, does not guarantee superior performance. 
Addressing this issue 
without relying on real-world, robot-damaging trajectories remains an unresolved challenge. 
In this work, we propose a grey-box model that progresses toward addressing this issue by combining: (i) the strength of black-box models to learn efficiently from real-world data, thus reducing the sim-to-real gap, and (ii) the strengths of white-box approaches to generalize to unseen data, thereby addressing the out-of-distribution problem.

\textbf{Our main contributions} are as follows.

\textbf{Neural-symbolic physics-aware layer}, which introduces a substantial physics prior that improves generalization on out-of-distribution terrains -- a crucial property for any reliable deployment.

\textbf{Explainability} of the proposed grey-box model provides several
well-interpretable intermediate outputs, such as contacts, slippage, or forces, that may serve as a natural source of self-supervision.

\textbf{Image-conditioned simulation:} The end-to-end differentiable image-conditioned simulation is suitable
for a myriad of robotics tasks such as control, learning or SLAM, as it predicts a vast number of trajectories in parallel.

\textbf{Experimental evaluation on non-rigid terrains:} The proposed model outperforms other state-of-the-art methods
especially on non-rigid terrains, such as grass or soft undergrowth that deforms when traversed by the robot. See Table~\ref{tab:results} for quantitative results. All codes and data are publicly available\footnote{\url{https://github.com/ctu-vras/fusionforce}}.
%In contrast to existing black-box models~\cite{Wellhausen-RAL-2019, hdif2023}, we argue that any well-generalizing model should be knowledgeable about the white-box laws of classical mechanics, such as (i) if the robot is falling, then it is accelerated by the gravity or (ii) any contact with the terrain results in forces acting on the robot body.

\begin{figure}
    \centering
    \includegraphics[width=0.5\textwidth]{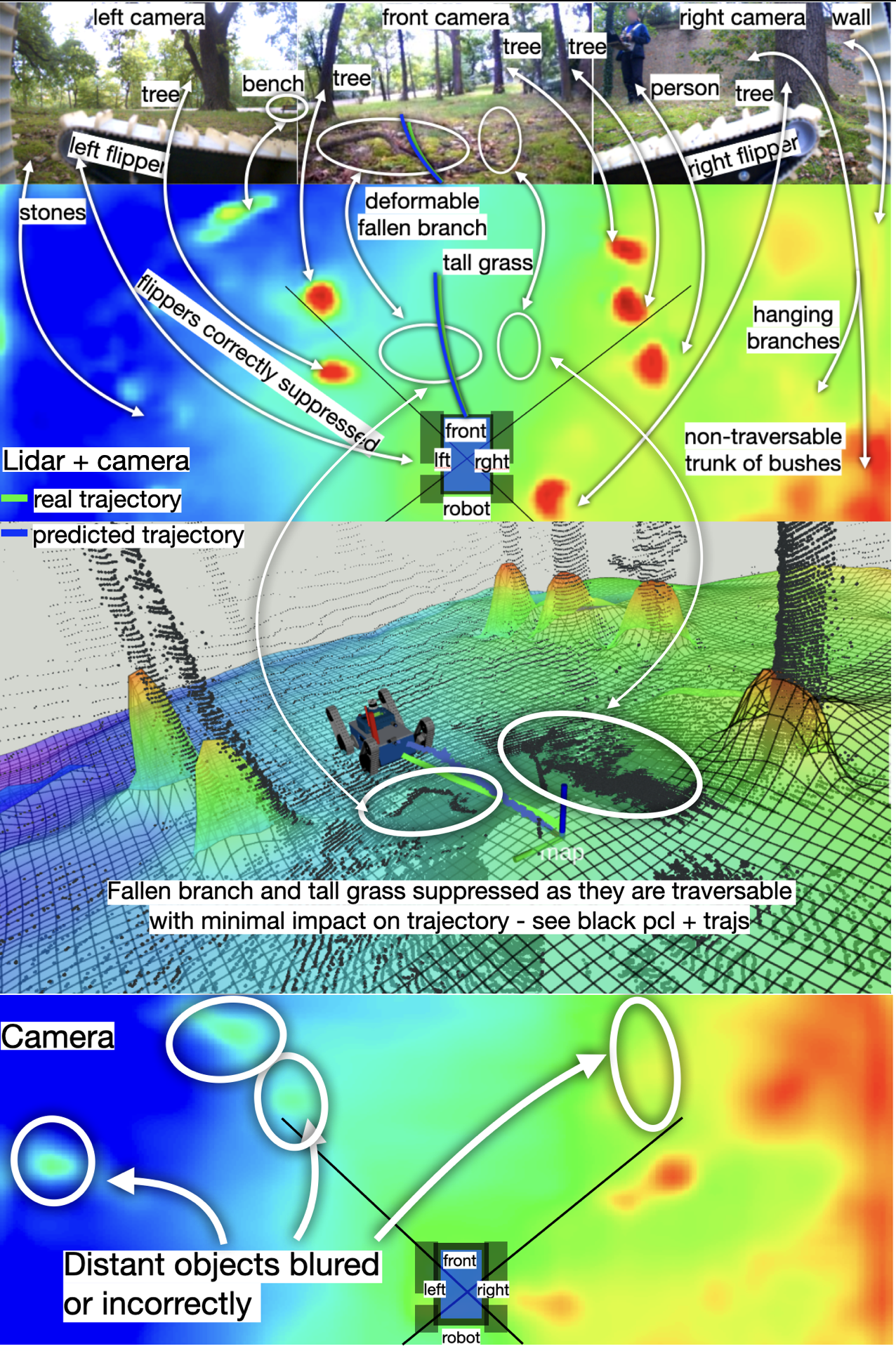}
    \caption{\textbf{Influence of lidar data} Input images (first row). The predicted supporting terrain from lidar + camera (second and third row) - see 3D pointcloud to compare accuracy. The predicted supporting terrain from camera only (last row).}
    \label{fig:terrain_encoders_results}
\end{figure}

\textbf{Relation to Previous Work:} A preliminary version of this work was presented at IROS 2024~\cite{agishev2024monoforce}. This journal submission substantially extends the prior work by:
\begin{itemize}
    \item Proposing novel architecture that allows to seamlessly integrate 2D and 3D sensor modalities, such as camera and lidar. It was shown that including the lidar data improves the accuracy of the estimation of terrain properties by 16\%, see Fig.~\ref{fig:terrain_encoders_results} for an example.

    \item Extending the predicted terrain properties by terrain friction,
    which significantly decreases the trajectory prediction error both in translation and in orientation (Table~\ref{tab:results}).
    See Fig.~\ref{fig:monoforce_heightmaps} for an example of predicted terrain properties.

    \item Improving the robot model for more accurate contact modeling and physical simulation.
    The new model is built from the robot's URDF and uses 223 robot body points in comparison to only 16 points used in the approximate robot model from~\cite{agishev2024monoforce}. See also qualitative results in Fig.~\ref{fig:catch-eye2}.

    \item Benchmarking against a purely data-driven baseline.
    It is shown that the proposed physics engine outperforms the LSTM-based baseline
    up to 56\% in translation and 48\% in orientation in terms of trajectory prediction accuracy, Fig.~\ref{fig:traj_errors}.
\end{itemize}

\section{Related work}
% \begin{figure}[t!]
% \centering
% %\includegraphics[width=0.5\textwidth]{imgs/results_generalization}
% \includegraphics[width=0.5\textwidth]{imgs/predictions/dphysics_vs_lstm/dphysics_lstm}
% \caption{\textbf{Generalization to robot-endangering scenarios} (non-present in dataset).
%     Qualitative example of trajectory predictions with $\nabla$Physics and TrajLSTM (data-driven) models.
%     The robot starts at the edge of a cliff and is given a command to move forward.
%     \textbf{Left}: robot-terrain interaction prediction with $\nabla$Physics model (a correct estimate of falling down).
%     \textbf{Right}: robot-terrain interaction prediction with TrajLSTM~\cite{pang2019aircraft} model
%     (incorrect prediction of hovering over a cliff as the robot-endangering situations are typically not present in datasets).}
%     \label{fig:dphysics_v_lstm}
% \end{figure}

Recent work has shown that learning can yield extremely robust locomotion controllers despite complex terrain dynamics. For example, Lee et al.~\cite{lee2020learning} train a neural controller for the ANYmal quadruped using only joint-encoder and IMU inputs, achieving robust zero-shot generalization to highly irregular outdoor terrain (mud, sand, vegetation, snow). Similarly, Choi et al.~\cite{choi2023learning} incorporate a parameterized physics model of granular media into the reinforcement learning loop, enabling a quadruped to run on soft sand and generalize to diverse surfaces (asphalt, grass, synthetic mat) by identifying substrate parameters on the fly. These works focus only on proprioceptive feedback and do not allow to predict possibly fatal behaviour from exteroceptive sensors such as camera or LiDAR. 

Complementary approaches use exteroceptive sensing to predict terrain traversability. Hoeller et al.~\cite{hoeller2023anymal} (ANYmal Parkour) employ onboard cameras and LiDAR to build a point-cloud map and latent terrain representation, guiding a high-level policy that selects from learned locomotion skills (walk, jump, climb, crouch) as the robot navigates obstacles. In wheeled off-road driving, Cai et al.~\cite{cai2022evora} (EVORA) train a deep model to predict distributions of wheel traction from a semantic–elevation map fused from images and LiDAR, and incorporate this into a model-predictive planner that avoids low-traction or high-uncertainty regions. Vecchio et al.~\cite{vecchio2023self} use self-supervised learning on synthetic RGB video to regress per-region traversability costs, bridging sim-to-real with domain adaptation so the robot can estimate navigational cost maps from raw camera images. These perceptive methods produce rich terrain-cost models from sensor data, but typically treat the mapping from visuals to robot state as a black box, with limited explicit physics.

A growing trend is to extend learned models with physical reasoning. For instance, the deformable-terrain work by Choi et al.~\cite{choi2023learning} embeds a physics model of sand within the controller, and EVORA explicitly models traction predictions to detect novel or out-of-distribution terrains. The AnyNav frameworkZ\cite{fu2025anynav} exemplifies this neuro-symbolic direction: it uses a neural network to predict a friction coefficient from camera images that is grounded by a symbolic friction model, and then a friction-aware planner chooses safe, physically-feasible paths. Such approaches improve generalization by enforcing consistency with physical laws. In summary, while recent literature has significantly advanced data-driven traversability estimation and control, a gap remains in merging perception with explicit physics. The proposed physics-aware model aims to fill this gap by learning interpretable terrain parameters (e.g. friction, stiffness see Fig.~\ref{fig:monoforce_heightmaps} for an example) from sensors and plugging them into neural-symbolic layer that delivers expected trajectories by solving the underlying ODE, thereby enhancing robustness on out-of-distribution terrains. Since our primary focus is on developing a reliable model for trajectory prediction, we discuss related works mainly from the perspective of terrain and trajectory modeling, while details of their specific use for control implementations are omitted.

\subsection{Black-box (Data-driven) Models}\label{subsec:black-box-models}

The modeling of environment representation from single~\cite{mani2020monolayout} or
multiple RGB images~\cite{philion2020lift} in a top-down view has been widely studied in the literature.
The geometry of the scene prediction (taking into account occlusions) from stereo or depth-camera input
has been addressed in~\cite{watson2020footprints}.
The importance and benefits of sensor fusion (camera images, lidar, and radar) for environment representation
have been demonstrated in~\cite{hendy2020fishing}.
The authors of RoadRunner~\cite{frey2024roadrunner} developed an end-to-end learning-based framework
that predicts the traversability and elevation maps from multiple images and a lidar voxel map.
However, the robot-terrain interaction and trajectory prediction tasks are not considered in these works.

The Recurrent Neural Networks (RNN) models~\cite{rumelhart1986learning} and
especially their variant Long Short-Term Memory (LSTM)~\cite{hochreiter1997long} and
Gated Recurrent Unit (GRU)~\cite{cho2014learning} have been widely and
successfully used~\cite{xie2020motion, yoon2022trajectory, pang2019aircraft}
for trajectory prediction tasks of various agents (mobile robots, cars, airplanes, people, and more).
However, their main drawback lies in the difficulty of capturing the underlying physical laws,
which leads to limited generalization capabilities
%(more in Section~\ref{subsec:data_driven_baseline} and Fig.~\ref{fig:dphysics_v_lstm}).

\begin{figure*}[th]
    \centering
    \includegraphics[width=\textwidth]{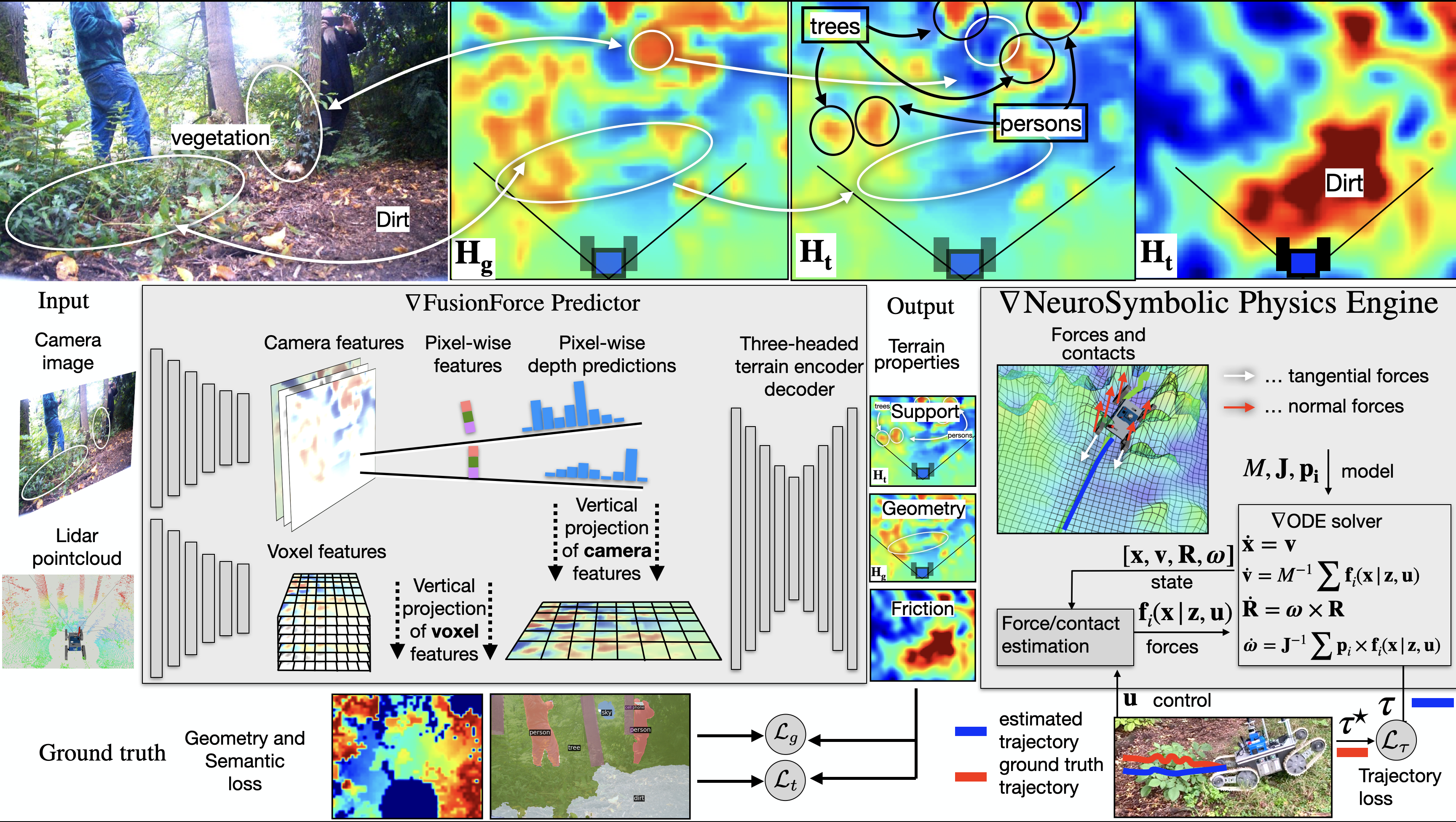}
    \caption{\textbf{Detailed architecture overview:} Our model consists of a data-driven \emph{FusionForce Predictor} and physics-driven differentiable \emph{Neuro-Symbolic Physics Engine}. The \emph{FusionForce Predictor} estimates rich features in image and voxel domains. Image features are then lifted to voxel domain according to the predicted pixel depths. Both camera a lidar features are vertically projected on the ground plane and three-headed terrain encoder-decoder is used to predict terrain properties.
    \emph{$\nabla$Neuro Symbolic Physics Engine} estimate forces at robot-terrain contacts by combining control signal, robot model and predicted properties. Resulting robot trajectory is integrated from these forces.
    Learning employs three losses: \emph{trajectory loss}, $\mathcal{L}_{\tau}$, which measures the distance between
    the predicted and real trajectory; \emph{geometrical loss}, $\mathcal{L}_g$, which measures the distance between
    the predicted geometrical heightmap and lidar-estimated heightmap;
    \emph{terrain loss}, $\mathcal{L}_t$, which enforces rigid terrain on rigid semantic classes revealed through image foundation model (SEEM~\cite{zou2023segment}).
    }
    \label{fig:monoforce}
\end{figure*}

\subsection{White-box (Physics-based) Models}\label{subsec:white-box-models}

The vehicle-terrain interaction modeling and traction mechanics have been widely studied in the literature~\cite{yong2012vehicle},~\cite{blau2008friction}.
To provide accurate analysis and prediction of off-road vehicles behavior on the terrain,
a set of high-fidelity physics-based software tools have been developed.
For example, Vortex Studio~\cite{vortexstudio2025} allows for realistic wheeled and tracked locomotion simulations
and extensive land and even planetary environment modeling.
The Chrono~\cite{serban2019integrated} is a physics-based engine designed for ground vehicle-tire-terrain interactions simulation.
It supports the integration of various vehicle and tire models and can simulate deformable terrain.
The AGX physics engine
(more specifically, its terrain dynamics model \textit{agxTerrain})~\cite{Berglund2019agxTerrain}
enables simulation of soil dynamics and interactions with heavy vehicles.
This software delivers realistic, high-fidelity simulations for applications
involving complex soil-tool interactions.
However, the mentioned physics engines are not differentiable,
which makes them unsuitable for end-to-end learning and integration of exteroceptive real sensor data.

\subsection{Grey-box (Hybrid) Models}\label{subsec:grey-box-models}

To achieve the accuracy and efficiency of data-driven methods and at the same time to
maintain the generalization capabilities of the physics-based models,
the \textit{hybrid} approaches have been developed for off-road navigation.
The PIAug~\cite{maheshwari2023piaug} is a physics-informed data augmentation methodology designed to enhance the learning
of vehicle dynamics, particularly in scenarios involving high-speed and aggressive maneuvers.
The robot's motion prediction in off-road scenarios is the main focus of the work, however,
the accuracy of terrain properties estimation is not addressed there.
The PhysORD~\cite{zhao2024physord} is a promising neuro-symbolic model;
the authors utilize neural networks to estimate the potential energy of a vehicle and
external forces acting on it during the terrain traversal.
The Hamiltonian and Lagrangian mechanic's laws are then used to predict the vehicle dynamics.
However, the integration of exteroceptive sensors (camera images and terrain properties)
is not implemented there, limiting the deployment of the method in real-world off-road scenarios.

%by supporting the integration of other sensor modalities, such as lidar (Section~\ref{subsec:sensor_fusion}), into the MonoForce model.
%It was shown that including the lidar data improves the accuracy of the terrain properties estimation by 15\% (Section~\ref{subsec:sensor_fusion}).
%Additionally, we use a precise robot model (built from the robot's URDF, the new model uses 223 robot body points
%comparing to only 16 points used in the approximate robot model from~\cite{agishev2024monoforce})
%for more accurate contact modeling and physical simulation compared to the previous work.
%It is demonstrated that this improvement decreases the trajectory prediction error
%by 14\% in translation and 26\% in orientation (Table~\ref{tab:results}).
%We also expand the data sequences (Section~\ref{subsec:rough_data}) in the newly introduced ROUGH dataset (Section~\ref{subsec:rough_data})
%that are used to train the model focusing on tracked robots and more challenging navigation scenarios
%(low traction, diverse terrain inclinations, etc)
%in comparison to the previously introduced RobinGas data~\cite{agishev2024monoforce}.
%Another contribution that builds on~\cite{agishev2024monoforce} is the performed benchmarking against
%a purely data-driven trajectory prediction baseline (Section~\ref{subsec:dphys_vs_lstm}).
%It is shown that the proposed physics engine outperforms the data-driven baseline
%up to 56\% in translation and 48\% in orientation in terms of trajectory prediction accuracy, Fig.~\ref{fig:traj_errors}.

\section{Methodology}
A detailed overview of the proposed architecture that converts images, pointclouds and control commands
into trajectories is depicted in~Fig.~\ref{fig:monoforce}.
The model consists of several learnable modules that deeply interact with each other.
The \emph{terrain encoder} carefully transforms visual features from the input image
into the heightmap domain using known camera geometry. These features are concatenated in the heightmap domain with voxel features estimated from lidar pointclouds.
The resultant heightmap features are further refined by the multihead encoder-decoder architecture into interpretable physical quantities
that capture properties of the terrain such as its shape, friction, stiffness, and damping.
Next, the \emph{neuro-symbolic physics engine} combines the terrain properties with the robot model,
robot state, and control commands and delivers reaction forces at points of robot-terrain contacts.
It then solves the equations of motion dynamics by integrating these forces
and delivers the trajectory of the robot.
Since the complete computational graph of the feedforward pass is retained,
the backpropagation from an arbitrary loss, constructed on top of delivered trajectories,
or any other intermediate outputs is at hand.

\subsection{Terrain Encoder}\label{subsec:terrain_encoder}

The part of the FusionForce architecture (Fig.~\ref{fig:monoforce})
that predicts terrain properties $\mathbf{m}$ from sensor measurements $\mathbf{z}$ is called \emph{terrain encoder}.
The proposed architecture starts by converting pixels from a 2D image plane into a heightmap with visual features.
Since the camera is calibrated, we leverage this substantial geometrical prior to connect heightmap bins with pixels.
%We incorporate the geometry through the Lift-Splat-Shoot architecture~\cite{philion2020lift}.
We use known camera intrinsic parameters \( \mathbf{K} \in \mathbb{R}^{3 \times 3} \) to estimate rays corresponding to particular pixels~--
pixel-wise rays.
For $i$-th ray, the convolutional network then predicts depth probabilities $p(d_i)$ and visual features $\Phi_i$. Given the depth $d_i$ in pixel $(u_i, v_i) $, we reconstruct the 3D position $\mathbf{P}_i$ as follows:
\begin{equation}
\mathbf{P}_i = d_i \cdot \mathbf{K}^{-1} 
\begin{bmatrix} u_i \\ v_i \\ 1 \end{bmatrix}, 
\end{equation}
therefore all visual features for all nonzero depth probabilities $p(d_i)>0$ can be vertically projected on a virtual heightmap. %for all depths along the corresponding ray that have nonzero probability $p(d_i)>0$.
Finally, the $p(d_i)$-weighted sum of visual features over each heightmap cell is computed. In particular, denoting $J$ the set of all 3D positions that are vertically projected into the heightmap cell $j$, the visual features $\varphi_j$ corresponding to this cell are computed as follows
\begin{equation}
\varphi_j = \frac{\sum_{i\in J}\Phi_i p(d_i)}{\sum_{i\in J} p(d_i)}. 
\end{equation}

Similarly, pointclouds are transformed into voxels and voxel-wise features are also vertically projected on heightmap domain.
The resulting multichannel array that appears as the concatenation of camera and lidar features in heightmap domain, is further refined by deep convolutional network with encoder-decoder architecture
to estimate the terrain properties $\mathbf{m}$ - physics quantities that influences the robot motion.

The terrain properties include the geometrical heightmap $\mathcal{H}_g$,
supporting terrain heightmap $\mathcal{H}_t$ (shape of the terrain layer hidden under the vegetation),
terrain friction map $\mathcal{M}$, soft-terrain map $\Delta \mathcal{H}$ (models a partially flexible layer of terrain (e.g. mud) that is hidden under flexible vegetation), stiffness map $\mathcal{K}$ (strength of the terrain to prevent penetration by the robot), and dampening map $\mathcal{D}$ (the last two are considered constant in the current implementation).

\subsection{Differentiable Neuro-Symbolic Physics Engine}\label{subsec:dphysics}
The differentiable physics engine solves the robot motion equation and estimates
the trajectory corresponding to the delivered forces.
The trajectory is defined as a sequence of robot states $\tau = \{s_0, s_1, \ldots, s_T\}$,
where $\mathbf{s}_t = [\mathbf{x}_t, \mathbf{v}_t, R_t, \boldsymbol{\omega}_t]$
is the robot state at time $t$,
$\mathbf{x}_t \in \mathbb{R}^3$ and $\mathbf{v}_t \in \mathbb{R}^3$ define the robot's position and linear velocity in the world frame,
$R_t \in \mathbb{R}^{3 \times 3}$ is the robot's orientation matrix, and $\boldsymbol{\omega}_t \in \mathbb{R}^3$ is the angular velocity.
To get the next state $\mathbf{s}_{t+1}$, in general, we need to solve the following ODE:
\begin{equation}
    \label{eq:state_propagation}
    \mathbf{\dot{s}}_{t+1} = f(\mathbf{s}_t, \mathbf{u}_t, \mathbf{z}_t)
\end{equation}
where $\mathbf{u}_t$ is the control input and $\mathbf{z}_t$ is the environment state.
In practice, however, it is not feasible to obtain the full environment state $\mathbf{z}_t$.
Instead, we utilize terrain properties $\mathbf{m}_t = [\mathcal{H}_t, \mathcal{K}_t, \mathcal{D}_t, \mathcal{M}_t]$
predicted by the terrain encoder.
In this case, the motion ODE~\eqref{eq:state_propagation} can be rewritten as:
\begin{equation}
    \label{eq:state_propagation_terrain}
    \mathbf{\dot{s}}_{t+1} = \hat{f}(\mathbf{s}_t, \mathbf{u}_t, \mathbf{m}_t)
\end{equation}

Let's now derive the equation describing the state propagation function $\hat{f}$.
The time index $t$ is omitted further for brevity.
We model the robot as a rigid body with total mass $m$ represented by a~set of mass points
$\mathcal{P} = \{(\mathbf{p}_i, m_i)\; | \; \mathbf{p}_i~\in~\mathbb{R}^3, m_i~\in~\mathbb{R}^+, i=1~\dots~N\}$,
where $\mathbf{p}_i$ denotes coordinates of the $i$-th 3D point in the robot's body frame.
We employ common 6DOF dynamics of a rigid body~\cite{contact_dynamics-2018} as follows:
\begin{equation}
  \begin{split}
    \dot{\mathbf{x}} &= \mathbf{v}\\
    \dot{\mathbf{v}} &= \frac{1}{m}\sum_i\mathbf{F}_i
  \end{split}
  \quad\quad
  \begin{split}
    \dot{R} &= \Omega R\\
    \dot{\boldsymbol{\omega}} &= \mathbf{J}^{-1}\sum_i \mathbf{p}_i\times\mathbf{F}_i
  \end{split}
  \label{eq:contact_dynamics}
\end{equation}
where $\Omega = [\boldsymbol{\omega}]_{\times}$ is the skew-symmetric matrix of $\boldsymbol{\omega}$.
We denote $\mathbf{F}_i\in\mathbb{R}^3$ a total external force acting on $i$-th robot's body point.
Total mass $m = \sum_i~m_i$ and moment of inertia $\mathbf{J}\in\mathbb{R}^{3\times 3}$ of the robot's rigid body are assumed to be known
static parameters since they can be identified independently in laboratory conditions.
Note that the proposed framework allows backpropagating the gradient with respect to these quantities, too,
which makes them jointly learnable with the rest of the architecture.
The trajectory of the rigid body is the iterative solution of differential equations~\eqref{eq:contact_dynamics},
that can be obtained by any ODE solver for given external forces and initial state (pose and velocities).

\begin{figure}[t]
    \centering
    \includegraphics[width=0.5\textwidth]{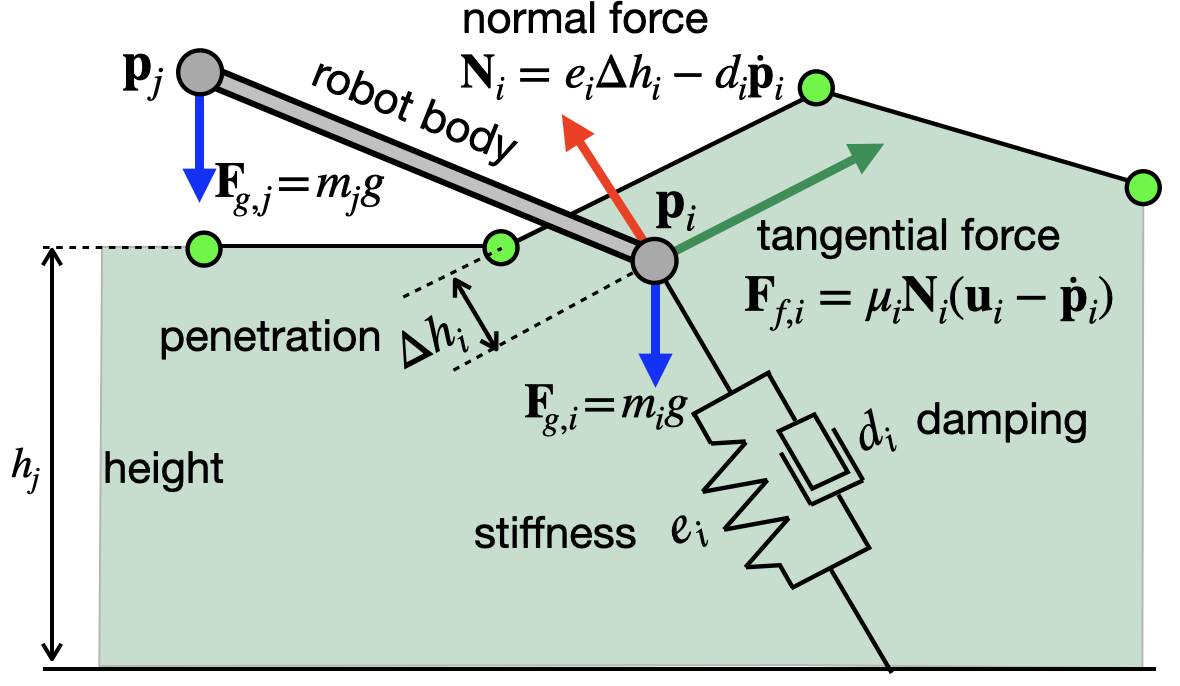}
    \caption{\textbf{Terrain force model}: Simplified 2D sketch demonstrating
    normal reaction forces acting on a robot body consisting of two points $p_i$ and $p_j$ .}
    \label{fig:spring_terrain_model}
\end{figure}

When the robot is moving over a terrain, two types of external forces are acting
on the point cloud $\mathcal{P}$ representing its model:
(i) gravitational forces and (ii) robot-terrain interaction forces.
The former is defined as $m_i\mathbf{g} = [0, 0, -m_ig]^\top$ and acts on
all the points of the robot at all times,
while the latter is the result of complex physical interactions that are not easy
to model explicitly and act only on the points of the robot that are in contact
with the terrain.
There are two types of robot-terrain interaction forces:
(i) normal terrain force that prevents the penetration of the terrain by the robot points,
(ii) tangential friction force that generates forward acceleration when the tracks are moving,
and prevents side slippage of the robot, see Fig.~\ref{fig:spring_terrain_model} for a simplified 2D sketch.

%\textbf{Robot-terrain interaction forces}

\noindent\textbf{Normal reaction forces:} 
One extreme option is to predict the 3D force vectors $\mathbf{F}_i$ directly
by a neural network, but we decided to enforce additional prior assumptions to reduce the risk of overfitting.
These prior assumptions are based on common intuition from the contact dynamics of flexible objects.
In particular, we assume that the magnitude of the force that the terrain exerts on the robot's point $\mathbf{p}_i\in \mathcal{P}$
increases proportionally to the deformation of the terrain.
Consequently, the network does not directly predict the force,
but rather predicts the height of the terrain $h\in\mathcal{H}_t$
at which the force begins to act on the robot body and the stiffness of the terrain $e\in\mathcal{K}$.
We understand the quantity $e$ as an equivalent of the spring constant from Hooke's spring model, \autoref{fig:spring_terrain_model}.
Given the stiffness of the terrain and the point of the robot that penetrated the terrain
by ${\Delta}h$, the reaction force is calculated as $e\cdot{\Delta}h$.
% \begin{figure}[t]
%     \centering
%     \includegraphics[width=0.4\textwidth]{imgs/dphysics/robot-terrain_forces}
%     \caption{\textbf{Robot-terrain interaction forces} acting on the robot's body at its contact points
%     with the terrain.
%     The point cloud was sampled from the MARV (\autoref{fig:robot_platforms}(b)) robot's 3D model.}
%     \label{fig:interaction_forces}
% \end{figure}

Since such a force, without any additional damping, would lead to an eternal bumping
of the robot on the terrain, we also introduce a robot-terrain damping coefficient $d\in\mathcal{D}$,
which similarly reduces the force proportionally to the velocity of the point
that is in contact with the terrain.
The model applies reaction forces in the normal direction $\mathbf{n}_i$ of the terrain surface,
where the $i$-th point is in contact with the terrain.
\begin{equation}\label{eq:normal_force}
    \mathbf{N}_{i} = \begin{cases}
 (e_i\Delta h_i - d_i(\dot{\mathbf{p}}_{i}^\top\mathbf{n}_i))\mathbf{n}_i  & \text{if } \mathbf{p}_{zi}\leq h_i \\
\mathbf{0} & \text{if } \mathbf{p}_{zi}> h_i
\end{cases},
\end{equation}
where terrain penetration $\Delta h_i = (h_i-\mathbf{p}_{zi})\mathbf{n}_{zi}$ is
estimated by projecting the vertical distance on the normal direction.
For a better gradient propagation, we use the smooth approximation of the Heaviside step function:
\begin{equation}
    \label{eq:smooth_normal_force}
    \mathbf{N}_i = (e_i\Delta h_i - d_i(\dot{\mathbf{p}}_{i}^\top\mathbf{n}_i))\mathbf{n}_i \cdot \sigma(h_i - \mathbf{p}_{zi}),
\end{equation}
where $\sigma(x) = \frac{1}{1+e^{-kx}}$ is the sigmoid function with a steepness hyperparameter $k$.

\begin{figure*}[t]
    \centering
    \includegraphics[width=\textwidth]{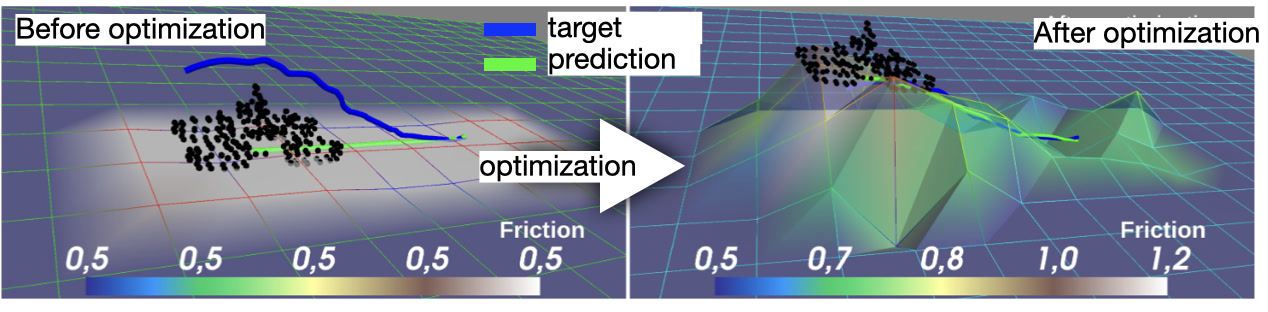}
    \caption{\textbf{Terrain computed by backpropagating through $\nabla$Physics:}
    Shape of the terrain (border of the area where terrain forces start to act) outlined by heightmap surface,
    its color represents the friction of the terrain.
    The optimized trajectory is in green, and the ground truth trajectory is in blue.}
    \label{fig:terrain_optim}
\end{figure*}

\noindent\textbf{Tangential friction forces:} 
Our tracked robot navigates by moving the main tracks and 4 flippers (auxiliary tracks).
The flipper motion is purely kinematic in our model.
This means that in a given time instant, their pose is uniquely determined by a $4$-dimensional vector
of their rotations, and they are treated as a rigid part of the robot.
The motion of the main tracks is transformed into forces tangential to the terrain.
The friction force delivers forward acceleration of the robot when robot tracks
(either on flippers or on main tracks) are moving.
At the same time, it prevents the robot from sliding sideways.
When a robot point $\mathbf{p}_i$, which belongs to a track, is in contact with terrain with
friction coefficient $\mu\in\mathcal{M}$, the resulting friction force at a contact point is computed as follows,~\cite{yong2012vehicle}:
\begin{equation}\label{eq:friction_force}
    \mathbf{F}_{f, i} = \mu_i |\mathbf{N}_i| ((\mathbf{u}_i - \mathbf{\dot{p}}_i)^\top\boldsymbol{\tau}_i)\boldsymbol{\tau}_i,
\end{equation}
where $\mathbf{u}_i = [u, 0, 0]^\top$, $u$ is the velocity of a track, and $\mathbf{\dot{p}}_i$ is the velocity of the point $\mathbf{p}_i$
with respect to the terrain transformed into the robot coordinate frame,
$\boldsymbol{\tau}_i$ is the unit vector tangential to the terrain surface at the point $\mathbf{p}_i$.
This model can be understood as a simplified Pacejka's tire-road model~\cite{pacejka-book-2012}
that is popular for modeling tire-road interactions.

To summarize, the state-propagation ODE~\eqref{eq:state_propagation_terrain}
(state $\mathbf{s}~=~[\mathbf{x},~\mathbf{v},~R,~\boldsymbol{\omega}]$) for a mobile robot moving over a terrain
is described by the equations of motion~\eqref{eq:contact_dynamics} where the force applied at a robot's $i$-th body point is computed as follows:
\begin{equation}\label{eq:forces}
    \begin{split}
        \mathbf{F}_i &= m_i\mathbf{g} + \mathbf{N}_i + \mathbf{F}_{f, i}
    \end{split}
\end{equation}
The robot-terrain interaction forces at contact points $\mathbf{N}_i$ and $\mathbf{F}_{f, i}$
are defined by the equations~\eqref{eq:smooth_normal_force} and~\eqref{eq:friction_force} respectively.

\noindent\textbf{Implementation of the Differentiable ODE Solver:} We implement the robot-terrain interaction ODE~\eqref{eq:contact_dynamics} in PyTorch~\cite{Paszke-NIPS-2019}.
The \textit{Neural ODE} framework~\cite{neural-ode-2021} is used to solve the system of ODEs.
For efficiency reasons, we utilize the Euler integrator for the ODE integration.
The differentiable ODE solver~\cite{neural-ode-2021} estimates the gradient through the implicit function theorem.
Additionally, we implement the ODE~\eqref{eq:contact_dynamics} solver that
estimates gradient through \textit{auto-differentiation}~\cite{Paszke-NIPS-2019},
i.e. it builds and retains the full computational graph of the feedforward integration.

\subsection{Data-driven Trajectory Prediction}\label{subsec:data_driven_baseline}
\begin{figure}
    \centering
    \includegraphics[width=0.5\textwidth]{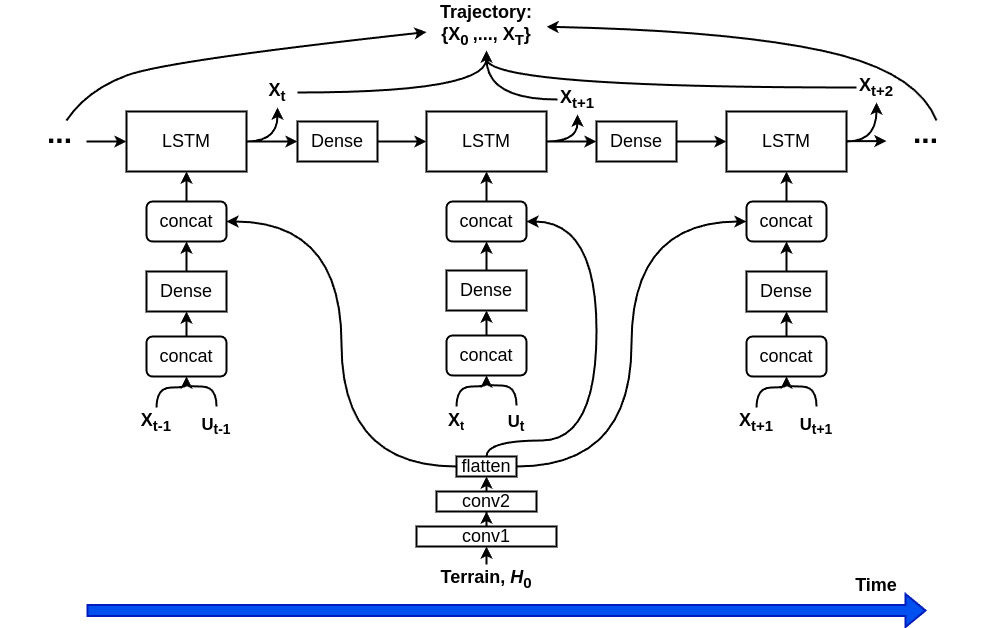}
    \caption{\textbf{TrajLSTM} architecture. The model takes as input: initial state $\mathbf{x}_0$, terrain $\mathcal{H}$, control sequence $\mathbf{u}_t, t \in \{0 \dots T\}$. It predicts the trajectory as a sequence of states $\mathbf{x}_t, t \in \{0 \dots T\}.$}
    \label{fig:traj_lstm}
\end{figure}

Inspired by the work~\cite{pang2019aircraft}, we design a data-driven LSTM architecture (Fig.~\ref{fig:traj_lstm}) for our outdoor mobile robot's trajectory prediction.
We call the model TrajLSTM and use it as a baseline for our $\nabla$Physics engine.
Given an initial robot's state $\mathbf{x}_0$ and a sequence of control inputs for a time horizon $T$, $\mathbf{u}_t, t \in \{0 \dots T\}$, the TrajLSTM model provides a sequence of states at control command time moments, $\mathbf{x}_t, t \in \{0 \dots T\}$.
As in outdoor scenarios the robot commonly traverses uneven terrain, we additionally include the terrain shape input to the model in the form of heightmap $\mathcal{H}=\mathcal{H}_0$ estimated at initial time moment $t=0$.
Each timestep's control input $\mathbf{u}_i$ is concatenated with the shared spatial features $\mathbf{x}_i$, as shown in Fig.~\ref{fig:traj_lstm}.
The combined features are passed through dense layers to prepare for temporal processing.
The LSTM unit~\cite{hochreiter1997long} processes the sequence of features (one for each timestep).
As in our experiments, the time horizon for trajectory prediction is reasonably small, $T=5 [\si{\sec}]$, and the robot's trajectories lie within the heightmap area, we use the shared heightmap input for all the LSTM units of the network.
So the heightmap is processed through the convolutional layers \textbf{once} and flattened, producing a fixed-size spatial feature vector.
This design choice (of not processing the heightmaps at different time moments) is also motivated by computational efficiency reason.
At each moment $t$, this heightmap vector is concatenated with the fused spatial-control features and processed by an LSTM unit.
The LSTM unit output for each timestep $t$ is passed through a fully connected (dense) layer to produce the next state $\mathbf{x}_{t+1}$.
The sequence of states form the predicted trajectory, $\{\mathbf{x}_0, \dots \mathbf{x}_T\}$.

\subsection{End-to-end Learning}\label{subsec:end2end_learning}
Self-supervised learning of the proposed architecture minimizes three different losses:

\textbf{Trajectory loss} that minimizes
the difference between SLAM-reconstructed trajectory $\tau^\star$ and predicted trajectory $\tau$:
\begin{equation}~\label{eq:traj_loss}
   \mathcal{L}_\tau = \|\tau-\tau^\star\|^2
\end{equation}

\textbf{Geometrical loss} that minimizes the difference between
ground truth lidar-reconstructed heightmap $\mathcal{H}_g^\star$
and predicted geometrical heightmap $\mathcal{H}_g$:
 \begin{equation}~\label{eq:geom_loss}
     \mathcal{L}_g = \|\mathbf{W}_g\circ(\mathcal{H}_g-\mathcal{H}_g^\star)\|^2
 \end{equation}
$\mathbf{W}_g$ denotes an array selecting the heightmap channel corresponding to the terrain shape.

\begin{figure}[t] 
    \centering
    \includegraphics[width=0.5\textwidth]{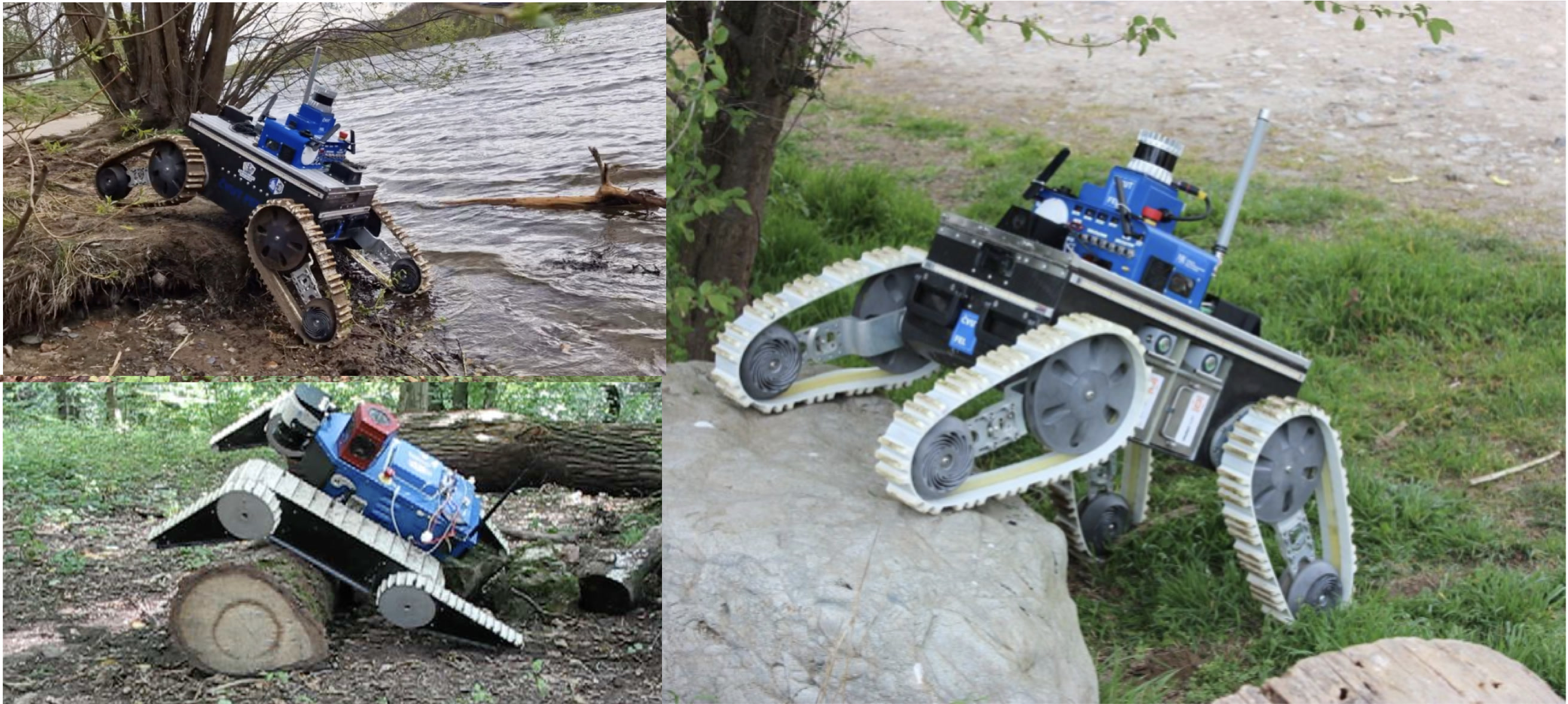}
%    \includegraphics[width=0.99\columnwidth]{imgs/robots/robots_blured}
    % \subfigure[\textit{Bluebotics Absolem} robot]{\includegraphics[width=0.38\columnwidth]{imgs/absolem}}
    % \subfigure[\textit{MARV} robot]{\includegraphics[width=0.44\columnwidth]{imgs/marv}}
    \caption{\textbf{Two robot platforms} used to collect the ROUGH dataset. Notice that one platform has only four massive flippers, and the other has also two main tracks.}
    \label{fig:robot_platforms}
\end{figure}

\textbf{Terrain loss} that minimizes the difference between ground truth $\mathcal{H}_t^\star$
and predicted $\mathcal{H}_t$ supporting heightmaps containing rigid objects detected
with Microsoft's image segmentation model SEEM~\cite{zou2023segment},
that is derived from Segment Anything foundation model~\cite{li2023semantic}:
 \begin{equation}~\label{eq:terrain_loss}
     \mathcal{L}_t = \|\mathbf{W}_t\circ(\mathcal{H}_t-\mathcal{H}_t^\star)\|^2
 \end{equation}
$\mathbf{W}_t$ denotes the array selecting heightmap cells that are covered by rigid materials
(e.g. stones, walls, trunks), and $\circ$ is element-wise multiplication.

Since the architecture Fig.~\ref{fig:model_overview} is end-to-end differentiable,
we can directly learn to predict all intermediate outputs just using trajectory loss~\eqref{eq:traj_loss}.
An example of terrain learning with the trajectory loss is visualized in Fig.~\ref{fig:terrain_optim}.
To make the training more efficient and the learned model explainable, we employ the
geometrical loss~\eqref{eq:geom_loss} and terrain loss~\eqref{eq:terrain_loss} as regularization terms.
stat

\begin{figure*}
    \centering
    \includegraphics[width=\textwidth]{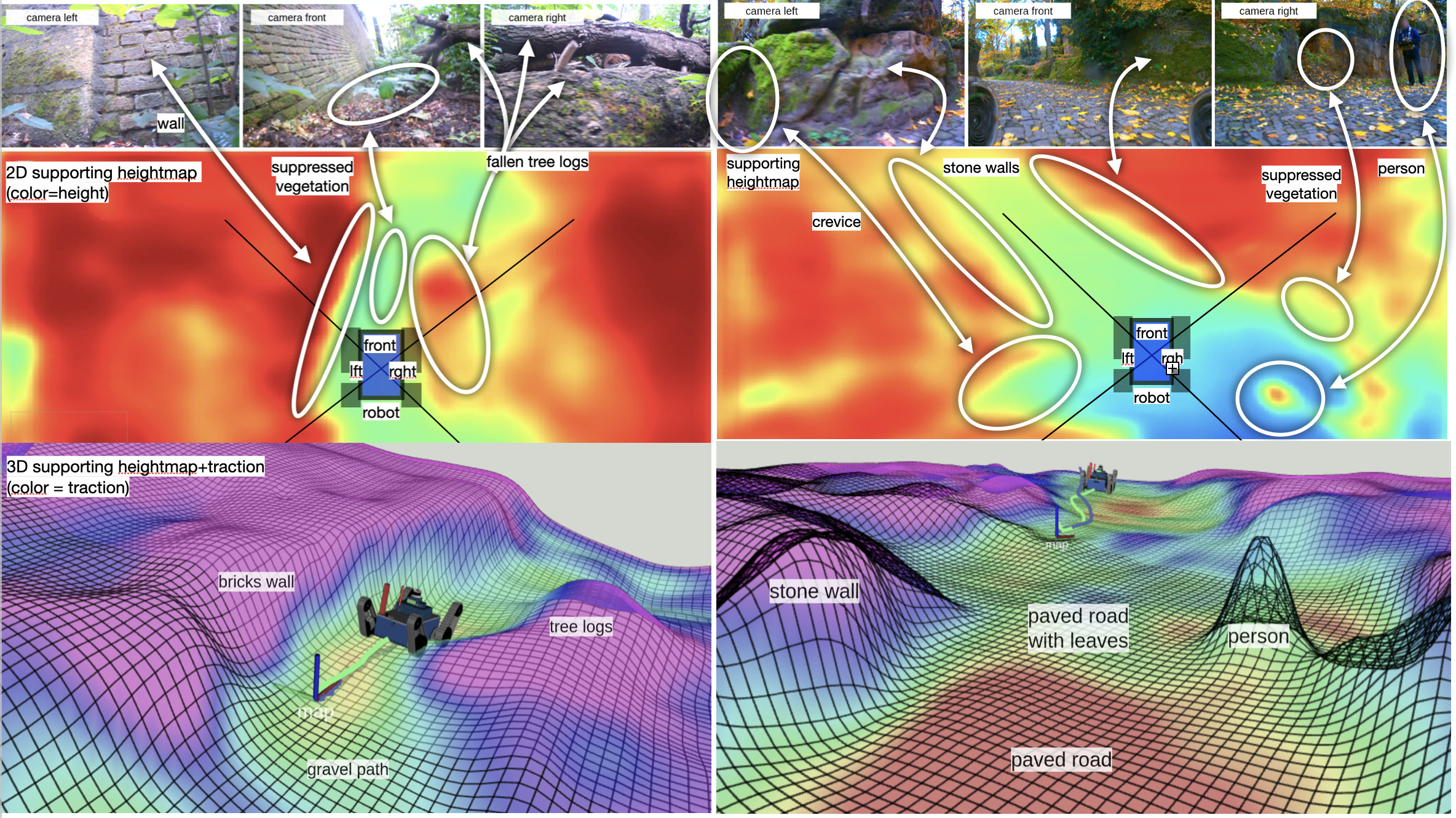}
    \caption{\textbf{Qualitative examples}.
    \emph{Left}: The robot is moving through a narrow passage between a wall and tree logs.
    \emph{Right}: The robot is moving on a gravel road with rocks on the sides.
    It starts its motion from the position marked with a coordinate frame and the trajectory is predicted for $10~[\si{\sec}]$ using real control commands.
    The camera images are taken from the robot's initial position (\emph{top row}).
    The visualization includes predicted supporting terrain $\mathcal{H}_t$ (\emph{second row}).
    It is additionally shown in 3D and colored with predicted friction values (\emph{third row}).
    }
    \label{fig:monoforce_predictions}
\end{figure*}

The Fig.~\ref{fig:monoforce_predictions} show the prediction examples of the FusionForce model in diverse outdoor environments.
From the example on the left,
we can see that the model correctly predicts the robot's trajectory and the terrain shape suppressing traversable vegetation,
while the rigid obstacles (wall and tree logs) are correctly detected.
The example on the right demonstrates the model's ability to predict the robot's trajectory ($10~[\si{\sec}]$-long horizon)
with reasonable accuracy and to detect the rigid obstacles (stones) on the terrain.
It could also be noticed that the surfaces that provide the robot good traction (paved and gravel roads) are marked with a higher friction value,
while for the objects that might not give good contact with the robot's tracks (walls and tree logs) the friction value is lower.

We argue that the friction estimates are approximate and an interesting research direction could be
comparing them with real-world measurements or with the values provided by a high-fidelity physics engine (e.g. AGX Dynamics~\cite{Berglund2019agxTerrain}).
However, one of the benefits of our differentiable approach is that the model does not require ground-truth friction values for training.
The predicted heightmap's size is $12.8\times12.8\si{\meter}^2$ and the grid resolution is $0.1\si{\meter}$.
It has an upper bound of $1~[\si{\meter}]$ and a lower bound of $-1~[\si{\meter}]$.
This constraint was introduced based on the robot's size and taking into account hanging objects (tree branches)
that should not be considered as obstacles (Fig.~\ref{fig:nav_monoforce}).
Additionally, the terrain is predicted in the gravity-aligned frame.
That is made possible thanks to the inclusion of camera intrinsics and extrinsics as input to the model,
Fig.~\ref{fig:monoforce}.
It also allows correctly modeling the robot-terrain interaction forces (and thus modeling the robot's trajectory accurately)
for the scenarios with non-flat terrain, for example, going uphill or downhill.
This will not be possible if only camera images are used as input.

\section{Results}
\begin{table*}
    \caption{Trajectory and terrain estimation accuracy. RGB = input is onboard camera image, PCL = input is onboard lidar pointcloud only, "hybrid" = architecture combining classical layers with the proposed neural symbolic physics engine, "data-driven" = architecture employing classical layers (feedforward and recurrent) only.}\label{tab:results}
    \centering
    \begin{tabular}{| c | c | c | c | c | c | c | c |}
    \hline
    input & method & terrain encoder & $\tau$ pred. & $\Delta \mathbf{x}$ [\si{\meter}] & $\Delta\mathbf{R}$ [\si{\mathrm{rad}}] & $\Delta \mathcal{H}_{g}$ [\si{\meter}] & $\Delta \mathcal{H}_{t}$ [\si{\meter}] \\
    \hline

    % RGB & hybrid & MonoForce~\cite{agishev2024monoforce} & $\nabla$Physics & \textbf{0.045} & \textbf{0.030} & 0.255 & 0.098 \\
    % \hline
%    RGB & hybrid & MonoForce~\cite{agishev2024monoforce} & $\nabla$Physics & 0.181 & 0.120 & 0.457 & 0.247 \\
%    \hline
    RGB & hybrid & MonoForce & $\nabla$Physics & 0.181 & 0.120 & 0.457 & 0.247 \\
    \hline

    RGB & hybrid & FusionForce & $\nabla$Physics & \textbf{0.043} & \textbf{0.035} & \textbf{0.193} & \textbf{0.074} \\
    \hline

     PCL & hybrid & FusionForce & $\nabla$Physics & \textbf{0.042} & \textbf{0.037}  & \textbf{0.156} & \textbf{0.064} \\
     \hline

    \makecell{RGB+PCL} & hybrid & FusionForce & $\nabla$Physics & \textbf{0.045} & \textbf{0.038} & \textbf{0.165} & \textbf{0.075} \\
    \hline 
    \hline

%    RGB & data-driven & LSS~\cite{philion2020lift} & TrajLSTM~\cite{pang2019aircraft}  & 0.128 & 0.070 &  \multicolumn{2}{c}{} \\
%    \cline{1-6}
    RGB & data-driven & LSS & TrajLSTM  & 0.128 & 0.070 &  \multicolumn{2}{c}{} \\
    \cline{1-6}

%    PCL & data-driven & VoxelNet~\cite{zhou2018voxelnet} & TrajLSTM~\cite{pang2019aircraft}  & 0.129 & 0.054 &  \multicolumn{2}{c}{} \\
%    \cline{1-6}
    PCL & data-driven & VoxelNet & TrajLSTM & 0.129 & 0.054 &  \multicolumn{2}{c}{} \\
    \cline{1-6}

%    \makecell{RGB+PCL} & data-driven & BEVFusion~\cite{liu2023bevfusion} & TrajLSTM~\cite{pang2019aircraft} & 0.125 & 0.053 & \multicolumn{2}{c}{} \\
%    \cline{1-6}
    \makecell{RGB+PCL} & data-driven & BEVFusion & TrajLSTM & 0.125 & 0.053 & \multicolumn{2}{c}{} \\
    \cline{1-6}

    \end{tabular}
\end{table*}

\subsection{Datasets}\label{subsec:rough_data}

We pre-train the terrain encoder on the large-scale outdoor dataset RELLIS-3D~\cite{jiang2020rellis3d}.
It is a multimodal dataset collected in an off-road environment containing accurately localized (with RTK GPS)
$13,556$ lidar scans and corresponding time-synchronized RGB images.

\begin{figure}[t!]
  \centering
  \includegraphics[width=0.5\textwidth]{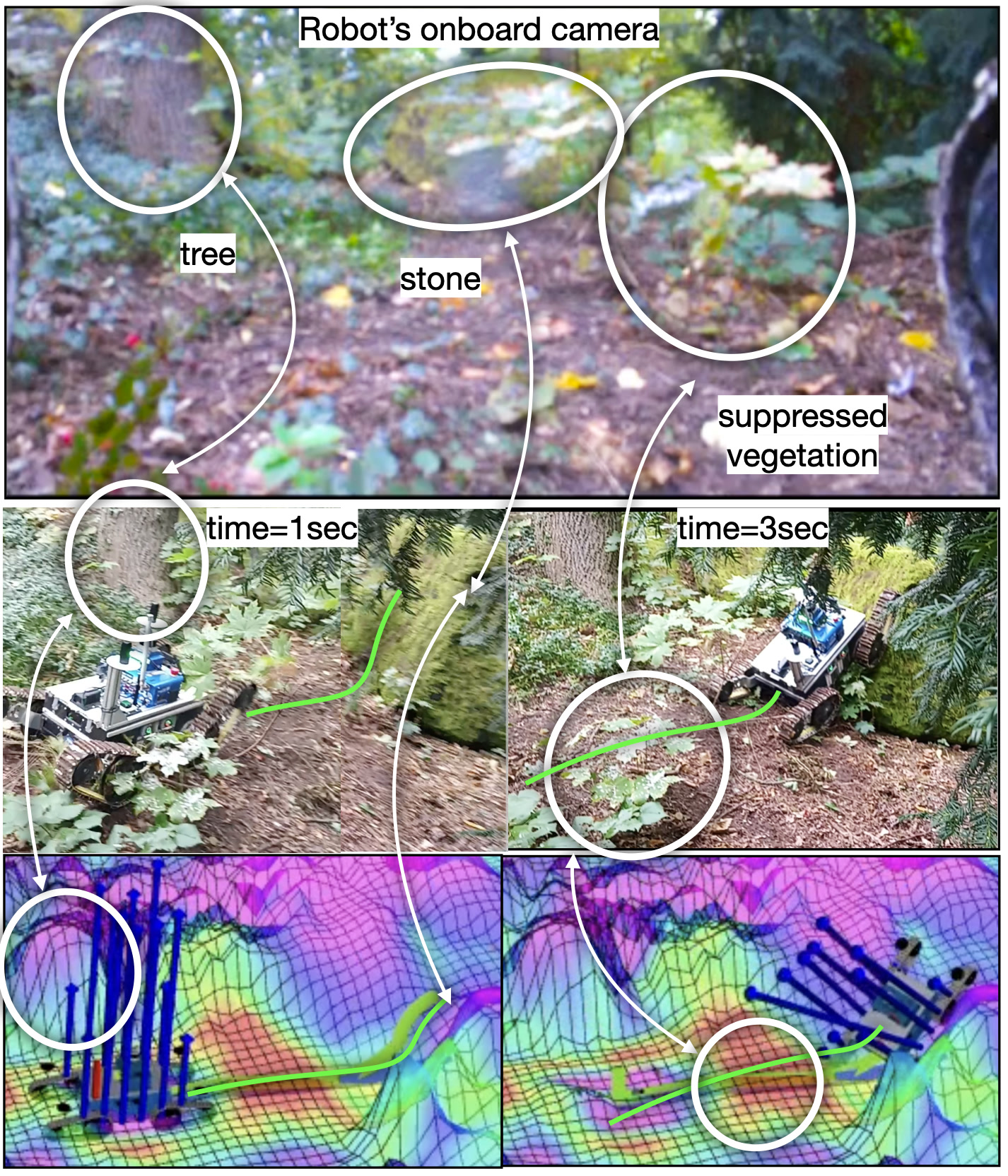}
  \caption{
  \textbf{Qualitative results:} Given input onboard image from the robot's camera (first row), the method distinguishes stones and trees as obstacles, while vegetation is suppressed (see 3D view in third row) the stone is preserved despite having almost the same green color as vegetation due to the moss coverage. Heightmap color encodes predicted terrain friction, and dark blue arrows show predicted contact forces. The predicted robot trajectory is visualized in light green. The ground truth robot motion is captured in the second row.}
  \label{fig:catch-eye2}
\end{figure}

Despite the amount and annotation quality of the data provided in the RELLIS-3D sequences, it lacks examples of a robot moving over hills, obstacles, and traversing high grass.
To fill this gap we manually collected several hours of driving with the flipper-based skid-steering robot in challenging terrain.
This dataset contains traversals on (or through) flexible natural obstacles like tall grass,
hanging tree branches, mud, dense undergrowth, see Figure~\ref{fig:catch-eye2}
for an example. Recorded traversal contains dynamically changing flipper positions to cover the variability of underlying physics problem. We refine RELIS-3D-pretrained FusionForce on these new data.

The rest of the experiments are organized as follows: Section~\ref{subsec:sensor_fusion} quantitatively compares proposed approach with respect to the other state-of-the-art approaches~\cite{philion2020lift,pang2019aircraft, zhou2018voxelnet,liu2023bevfusion}, including our own previous work MonoForce~\cite{agishev2024monoforce}; Section~\ref{subsec:dphys_vs_lstm} demonstrate superiority of the proposed $\nabla$Physics layer approach when compared to fully-data-driven approach.
Section~\ref{subsec:computational_efficiency} summarize computation efficiency of the proposed $\nabla$Physics layer on different hardware; Section~\ref{subsec:navigation} summarize results from autonomous navigation with FusionForce and MPPI in rough outdoor environment.

\subsection{Sensor Fusion}\label{subsec:sensor_fusion}

We train and evaluate the proposed FusionForce model with the four state-of-the-art baseline architectures. The first three considered architectures consist of LSS~\cite{philion2020lift}, VoxelNet~\cite{zhou2018voxelnet},
and BEVFusion~\cite{liu2023bevfusion} used as terrain encoders, followed by a data-driven trajectory predictor LSTM~\cite{pang2019aircraft}. We refer to these architectures as \emph{data-driven}. The last considered baseline is our own previous work called MonoForce~\cite{agishev2024monoforce} that uses $\nabla$-physics neuro symbolic layer for trajectory prediction. We refer to this architecture as \emph{hybrid}. All networks are trained using the losses described in Section~\ref{subsec:end2end_learning} on the same training and evaluation data. 

All approaches are compared in terms of two different metrics: (i) accuracy of trajectory prediction and (ii) accuracy of terrain properties prediction. The former represents accuracy computed on real trajectories, which is inherently limited only to safe traversals/terrains, while the latter mostly represents accuracy on dangerous terrains. Consequently, both metrics are equally important. Table~\ref{tab:results} summarizes the results. 

The trajectory prediction accuracy is reported as translational and rotational error computed as follows:
\begin{equation}~\label{eq:xyz_diff}
    \Delta\mathbf{x}=\sqrt{\frac{1}{T}\sum_{t=0}^T \|\mathbf{x}_t - \mathbf{x}_t^{*}\|}
\end{equation}
\begin{equation}~\label{eq:rot_diff}
    \Delta\mathbf{R}=\frac{1}{T}\sum_{t=0}^{T}\arccos\frac{\mathrm{tr}({\mathbf{R}^{\top}_t\mathbf{R}^{*}_t})-1}{2},
\end{equation}
 where $\{\mathbf{x}_t$ is predicted translation, $\mathbf{R}_t\}$ rotation and $\{\mathbf{x}_t^{*}, \mathbf{R}_t^{*}\}$, $t \in \{0 \dots T\}$ is the ground truth translation and rotation.
The ground truth poses were recorded using the SLAM method introduced in~\cite{Pomerleau-2013-AR}.

Similarly, the terrain prediction accuracy is evaluated as the $L2$-error of geometrical $\Delta\mathcal{H}_g$~(\ref{eq:geom_loss}),
and terrain $\Delta\mathcal{H}_t$~(\ref{eq:terrain_loss}) heightmaps on evaluation data. 
These metrics are shown in the two right columns of the Table~\ref{tab:results}.
We observe that FusionForce achieves, independently of the employed input modality, consistently smaller terrain prediction and trajectory estimation errors compared to the MonoForce baseline.
The improvement in terrain accuracy suggests a better generalization on dangerous (out-of-distribution) terrains.
% We observe that FusionForce achieves, independently of the employed input modality, consistently smaller terrain prediction errors compared to the MonoForce baseline, while the accuracy of the trajectory prediction is preserved. We conclude that there is no improvement in trajectory prediction because the accuracy of MonoForce on safe trajectories is already high. However, the improvement in terrain accuracy suggests a better generalization on dangerous (out-of-distribution) terrains.

The qualitative results of the prediction of terrain properties are visualized in Fig.~\ref{fig:terrain_encoders_results}.
Although the camera-only model provides a good estimate of the terrain shape
(and thus the predicted trajectory matches the ground truth quite closely), it struggles to predict obstacles with sharp edges such as tree trunks.
Using the lidar pointcloud as input provides more accurate terrain predictions.
%The BEVFusion prediction is also visualized in 3D in Fig.~\ref{fig:catch-eye} with the lidar point cloud input.
% \begin{table*}
%     \caption{Trajectory and terrain estimation accuracy.}\label{tab:results}
%     \centering
%     \begin{tabular}{c | c | c | c | c | c | c | c}
%     \hline
%     input & method & terrain encoder & $\tau$ pred. & $\Delta \mathbf{x}$ [\si{\meter}] & $\Delta\mathbf{R}$ [\si{\deg}] & $\Delta \mathcal{H}_{g}$ [\si{\meter}] & $\Delta \mathcal{H}_{t}$ [\si{\meter}] \\
%     \hline

%     RGB & hybrid & MonoForce~\cite{agishev2024monoforce} & $\nabla$Physics & 0.045 & \textbf{1.712} & 0.255 & 0.098 \\
%     \hline

%     RGB & hybrid & FusionForce & $\nabla$Physics & 0.043 & 2.013 & \multirow{2}{*}{0.193} & \multirow{2}{*}{0.074} \\
%     \cline{1-6}

%     RGB & data-driven & LSS~\cite{philion2020lift} & TrajLSTM~\cite{pang2019aircraft}  & 0.128 & 3.949 &  &  \\
%     \hline

%     point cloud & hybrid & FusionForce & $\nabla$Physics & \textbf{0.042} & 2.123  & \multirow{2}{*}{\textbf{0.156}} & \multirow{2}{*}{\textbf{0.064}} \\
%     \cline{1-6}

%     point cloud & data-driven & VoxelNet~\cite{zhou2018voxelnet} & TrajLSTM~\cite{pang2019aircraft}  & 0.129 & 3.369 &  & \\
%     \hline

%     \makecell{RGB +\\ point cloud} & hybrid & FusionForce & $\nabla$Physics & 0.045 & 2.254 & \multirow{3}{*}{0.165} & \multirow{3}{*}{0.075} \\
%     \cline{1-6}

%     \makecell{RGB +\\ point cloud} & data-driven & BEVFusion~\cite{liu2023bevfusion} & TrajLSTM~\cite{pang2019aircraft} & 0.125 & 3.067 & & \\
%     \hline

%     \end{tabular}
%\end{table*}

\subsection{Physics-based vs Data-driven Baseline}\label{subsec:dphys_vs_lstm}

The following case study compares our hybrid approach, which combines classical layers with a neural symbolic physics engine layer,
to a purely data-driven method in terms of trajectory estimation accuracy. 
We compare the physics model $\nabla$Physics with the data-driven TrajLSTM network in terms of trajectory prediction accuracy.

%(Section~\ref{subsec:dphysics})(Section~\ref{subsec:data_driven_baseline})

\begin{figure}
    \centering
    % \subfigure{\includegraphics[width=0.5\textwidth]{imgs/predictions/dphysics_vs_lstm/prediction_errors}}
    \subfigure{\includegraphics[width=0.49\columnwidth]{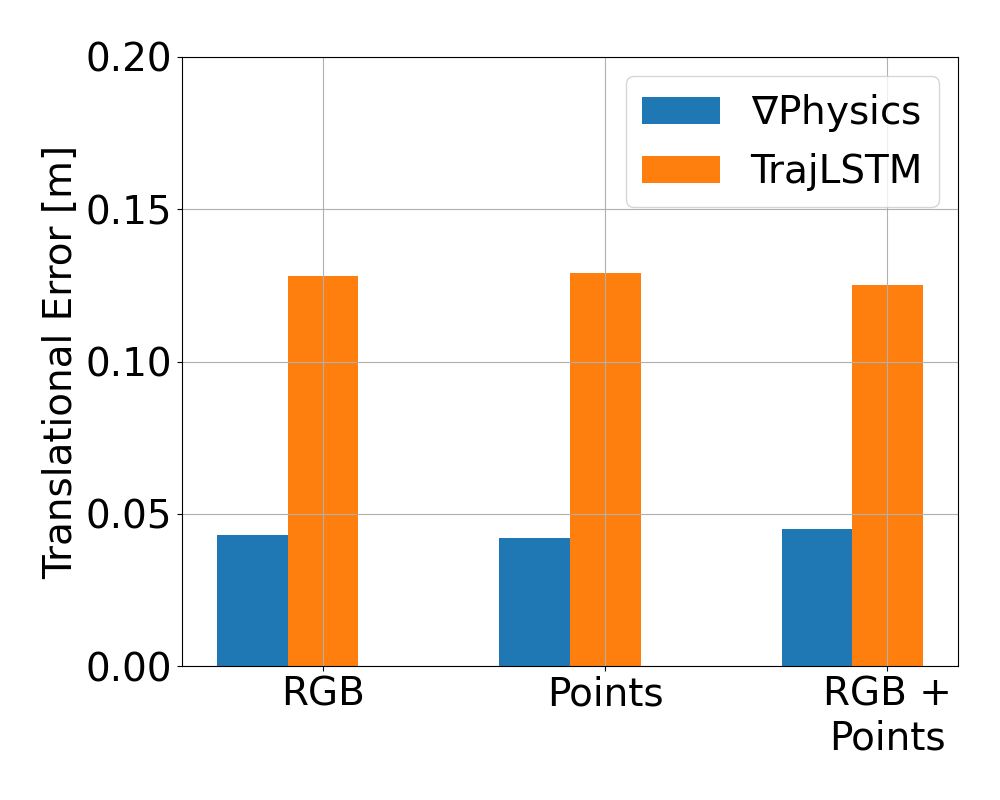}}
    \subfigure{\includegraphics[width=0.49\columnwidth]{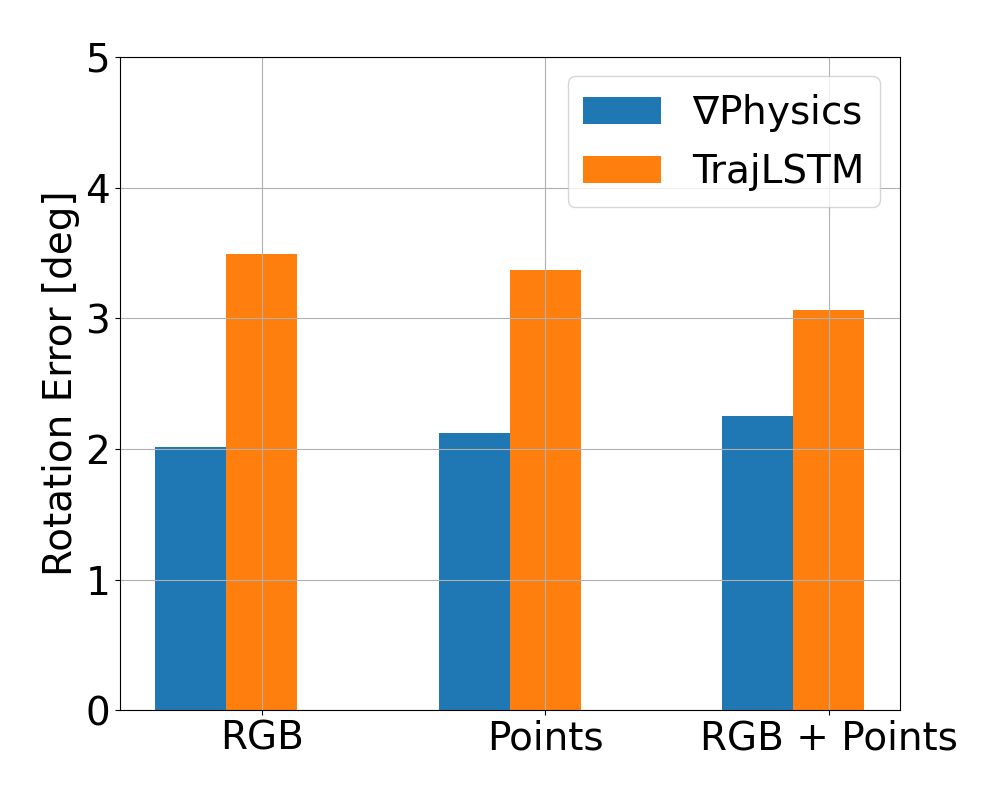}}
    \caption{\textbf{Generalization and results:} The proposed neuro-symbolic layer $\nabla$Physics generalizes to out-of-distribution examples well.
    The overall results are consistently better independent of the architecture
        (LSS~\cite{philion2020lift}, VoxelNet~\cite{zhou2018voxelnet}, BEV-Fusion~\cite{liu2023bevfusion}) used for terrain properties estimation.}
    \label{fig:traj_errors}
\end{figure}

The results are summarized in Table~\ref{tab:results} (columns $\Delta\mathbf{x}$ and $\Delta\mathbf{R}$)
and Fig.~\ref{fig:traj_errors}.
Note that for the fair comparison, the TrajLSTM architecture is designed in a way that
it has the same interface as the $\nabla$Physics module (takes the same input and yields the same output).
This allows to use of the $\nabla$Physics and TrajLSTM models interchangeably
with different terrain encoder models
(LSS~\cite{philion2020lift}, VoxelNet~\cite{zhou2018voxelnet}, BEVFusion~\cite{liu2023bevfusion})
that provide terrain estimates for the trajectory predictors.
It can be seen from Fig.~\ref{fig:traj_errors} that the physics-driven
trajectory predictor ($\nabla$Physics) provides better trajectory prediction
(both in translational and rotational components) regardless of the terrain estimation method.
Additionally, for the data-driven trajectory predictor (TrajLSTM) we observe an impact of sensor fusion.
The usage of BEVFusion terrain encoder helps to better estimate the translation and rotation of the robot
w.r.t. its single-modality baselines (image-based LSS and lidar-based VoxelNet).

Overall, the point cloud input helps to predict terrain more accurately
(Section~\ref{subsec:sensor_fusion} and Table~\ref{tab:results})
and the physics-based trajectory predictor $\nabla$Physics provides better results
than its data-driven baseline TrajLSTM (Fig.~\ref{fig:traj_errors}).

\subsection{Physics Engine Computational Efficiency}\label{subsec:computational_efficiency}
\begin{figure}
    \centering
    \includegraphics[width=0.48\textwidth]{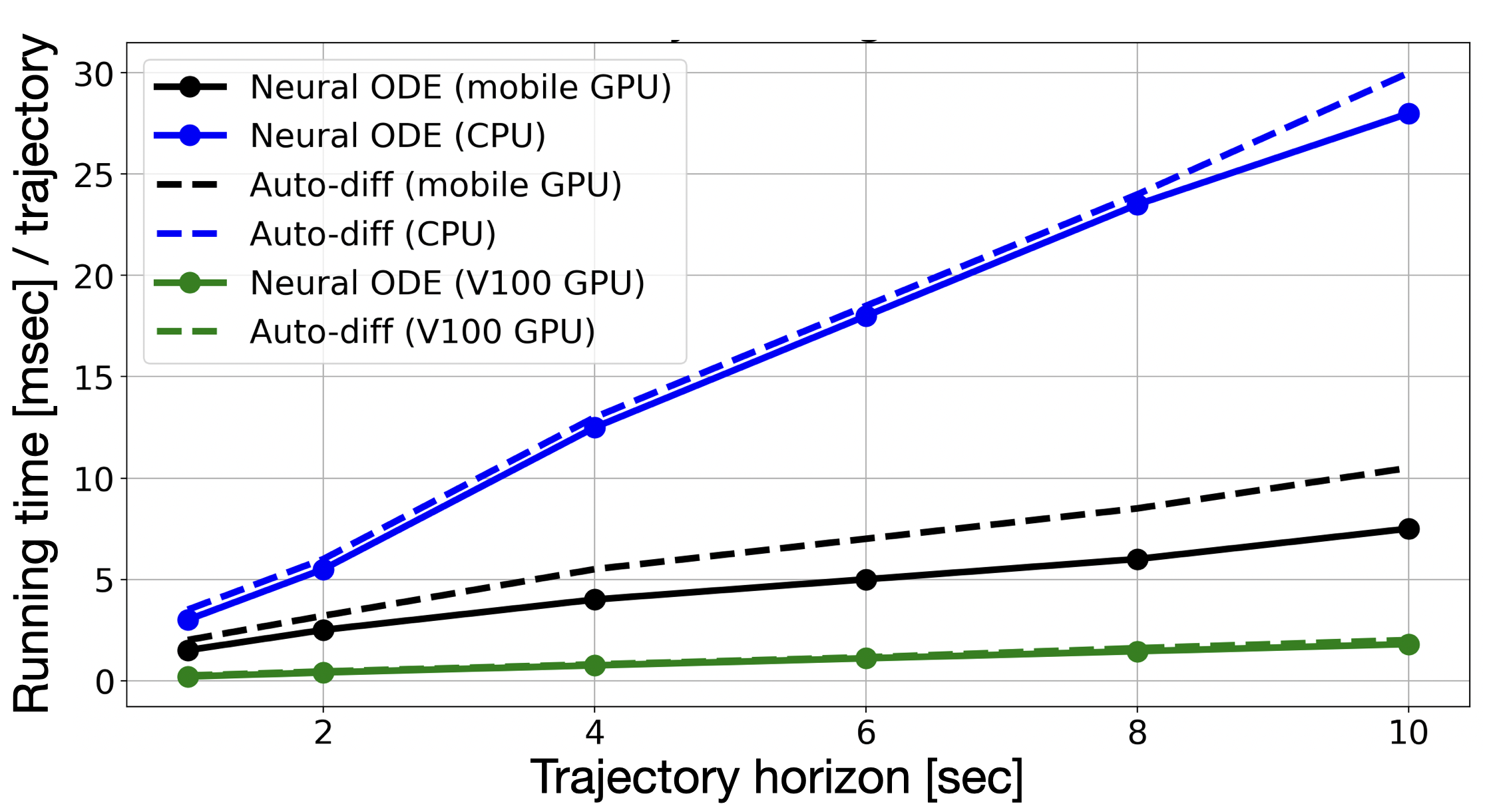}
    % \subfigure[\emph{CPU}:~AMD Ryzen 7 4800H, \emph{GPU}:~GeForce GTX 1660 Ti Mobile, $N_{traj}=512$]{\includegraphics[width=0.48\textwidth]{imgs/results/dphys_runtime_ntrajs_512_laptop}}
    % \subfigure[\emph{GPU}:~Tesla V100-SXM2-32GB, $N_{traj}=2048$]{\includegraphics[width=0.48\textwidth]{imgs/results/dphys_runtime_ntrajs_2048_V100_GPU}}

    \caption{The results are provided with the following configurations:
    grid reslution: $0.1~[m]$, number of robot body points: $223$ (uniformly sampled with voxel size $0.1~[m]$).}
    \label{fig:dphys_runtime}
\end{figure}
The overall learning behavior (speed and convergence) is mainly determined by
(i) the time horizon over which the difference between trajectories is optimized and
(ii) the way the differentiable ODE solver is implemented.
We compare further the computational efficiency of the two implementations
of the differentiable physics engine: based on \textit{Neural ODE} and \textit{auto-differentiation}.

\begin{figure*}
    \centering
    \includegraphics[width=\textwidth]{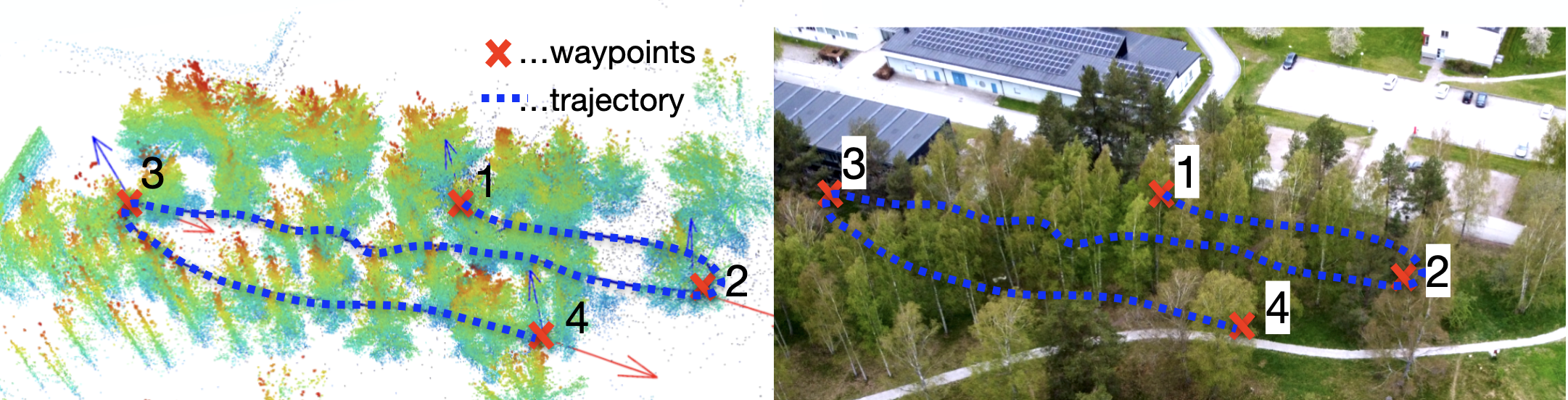}
    % \subfigure[The constructed point cloud map of the forest environment and robot's path during the navigation experiment;
    % the coordinate frames denote waypoints locations]{\includegraphics[width=0.40\textwidth]{imgs/navigation/pcd}}
    % \subfigure[The aerial photo of the forest area was taken for reference]{\includegraphics[width=0.45\textwidth]{imgs/navigation/drone_map}}
    \caption{The top-down view of the navigation experiment in the forest environment.
    During the experiment, the robot autonomously traverses the 260-meter-long path.
    The point cloud map construction~(a) and robot's localization were performed using the ICP SLAM method~\cite{Pomerleau-2013-AR}}.
    \label{fig:nav_maps}
\end{figure*}
\begin{figure*}
    \centering
    \includegraphics[width=\textwidth]{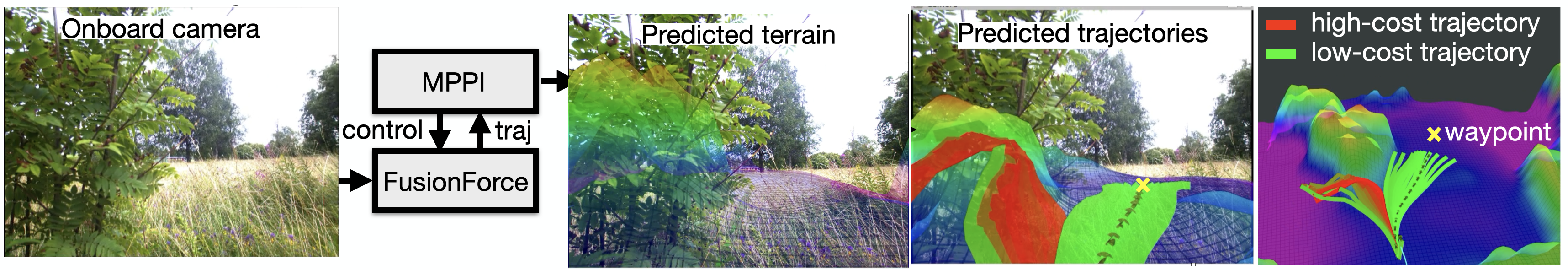}

    \subfigure[Going between tree trunks]{\includegraphics[width=0.24\textwidth]{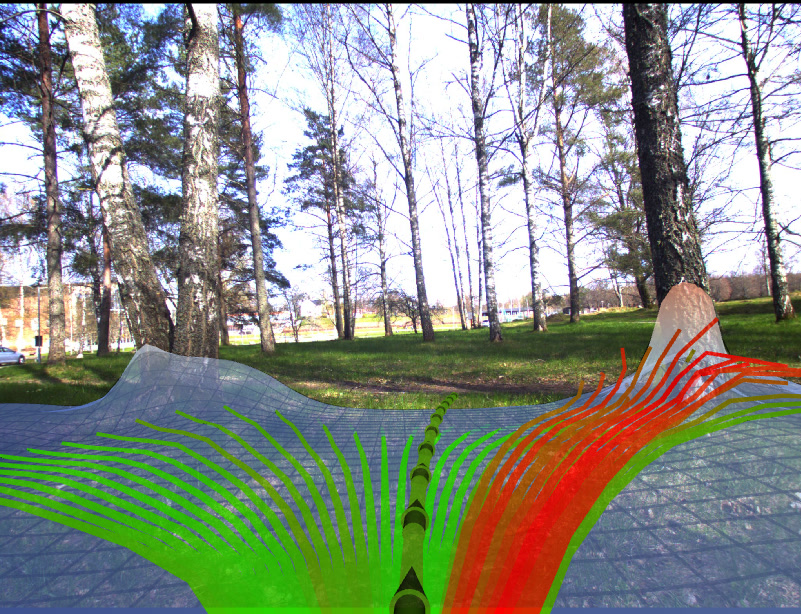}}
    \subfigure[Traversing hanging brunches]{\includegraphics[width=0.24\textwidth]{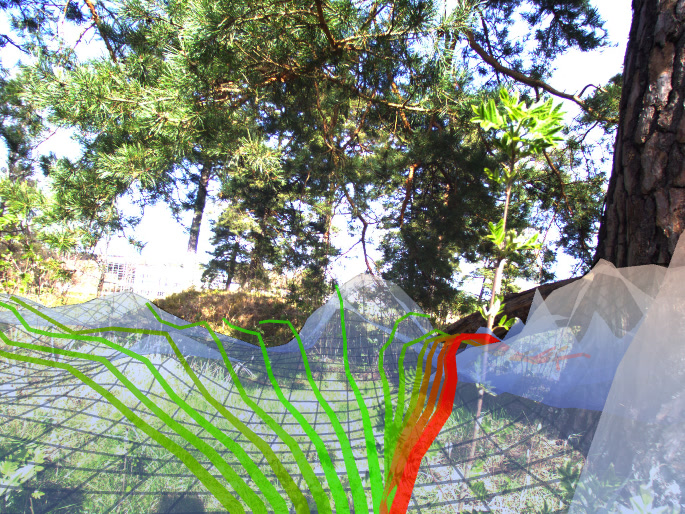}}
    \subfigure[Suppressing fallen brunches]{\includegraphics[width=0.24\textwidth]{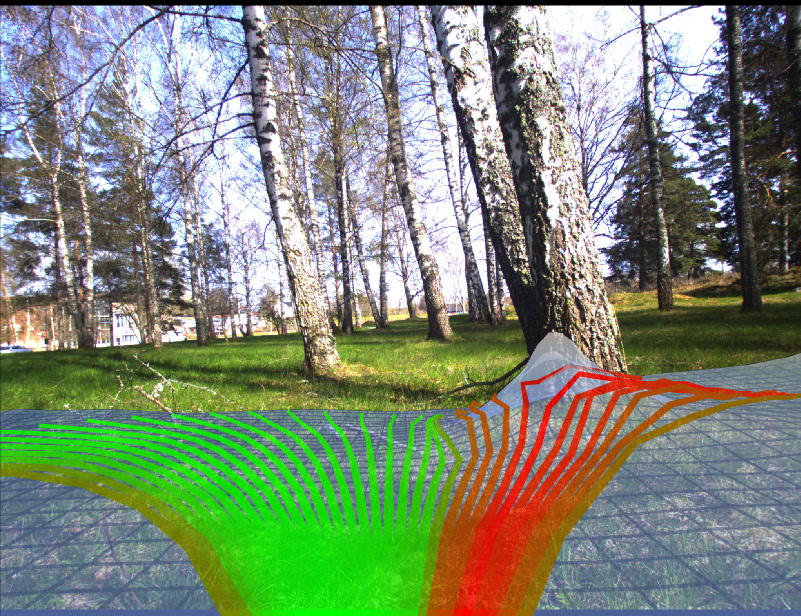}}
    \subfigure[Detecting stones in high vegetation]{\includegraphics[width=0.24\textwidth]{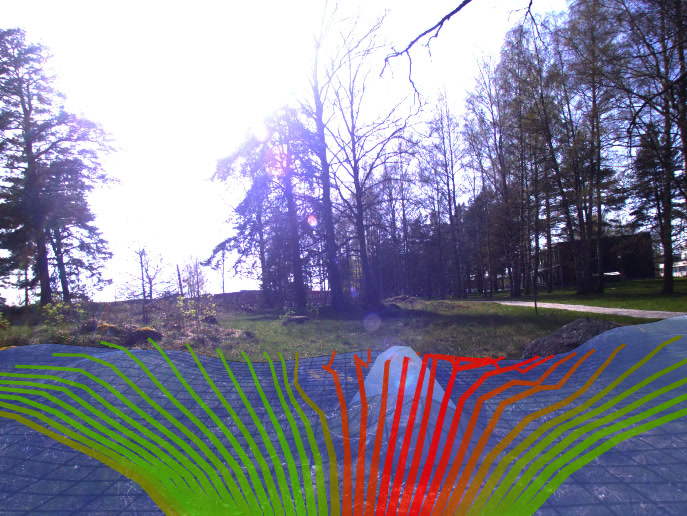}}

    \caption{Autonomous navigation in the forest environment.
    The robot follows a set of waypoints and at the same time avoids obstacles (trees, bushes, rocks, etc.).
    \textbf{First row (Control setup):}
    Trajectory predictions, based only on the onboard camera, for different control commands are sampled through a simplified MPPI technique.
    The colors of a possible trajectory correspond to the cost (red is the most expensive), and the yellow cross denotes the position of the closest waypoint.
    The trajectory with green arrows is the one with the smallest total cost,~\eqref{eq:total_cost}.
    \textbf{Second row (Qualitative results)}: Predicted supporting terrain and robot trajectory given camera images.
    The predictions and frontal camera images correspond to different time moments of the navigation experiment in the forest environment~(Fig.~\ref{fig:nav_maps}).
    % \textbf{Left}: view from the robot's frontal camera.
    % \textbf{Right}: MonoForce prediction visualized in 3D for the frontal camera field of view.
    }
    \label{fig:nav_monoforce}
\end{figure*}

The Fig.~\ref{fig:dphys_runtime} shows the runtime of the differentiable physics engine depending on
the predicted trajectories time horizon for the two solvers' implementation methods and different hardware configurations.
%The Fig.~\ref{fig:dphys_runtime}~(a) shows the runtime on the standard laptop hardware (CPU and GPU),
%while the Fig.~\ref{fig:dphys_runtime}~(b) shows the runtime on the high-performance GPU.
It can be noticed that the runtime grows linearly with the time horizon in all cases and
the V100 GPU implementation is around 10 times faster than the CPU one.
The \textit{Neural ODE} solver~\cite{neural-ode-2021} is slightly more efficient than the \textit{auto-differentiation} one.
For example, on a high-performance GPU (Tesla V100-SXM2-32GB), the prediction of
2048 trajectories ($6~[\si{\sec}]$-long) takes around $1~[\si{\sec}]$.
The short runtime of the differentiable physics engine allows for its
efficient usage in the end-to-end learning pipeline %(Section~\ref{subsec:end2end_learning}) 
and for real-time
trajectory shooting in navigation scenarios. %(Section~\ref{subsec:navigation}).

\subsection{Autonomous Navigation in Unstructured Environments}\label{subsec:navigation}

% \begin{figure}
%     \centering
%     \includegraphics[width=0.5\textwidth]{imgs/navigation/petrin/monoforce_petrin}
%     \caption{Prediction projected to robot's camera images.
%     The prediction includes supporting terrain $H_t$;
%     the colors correspond to the friction coefficient $\mu$ (violet is low, red is high).
%     A set of 64 trajectories for different control commands is visualized in white and black (low and high trajectory cost~\eqref{eq:traj_cost} respectively).
%     The black frames denote the camera's field of view.}
%     \label{fig:nav_petrin}
% \end{figure}
We demonstrate that the proposed model can be easily used for autonomous navigation in unstructured outdoor environments.
In the navigation experiments, the robot is given a set of distant waypoints to follow and it is localized using the ICP SLAM method~\cite{Pomerleau-2013-AR}.
The model runs onboard the robot's hardware and predicts terrain properties in front of the robot and a set of possible trajectories for different control commands.
In the case of a tracked robot platform (Fig.~\ref{fig:robot_platforms}), the controls are the commanded velocities of individual tracks.
Sample of predicted trajectories for given controls is visualized in Fig.~\ref{fig:nav_monoforce}.
To reach a goal safely, the robot chooses the trajectory with the smallest cost and distance to the next waypoint.

Thanks to the ability of the FusionForce to predict robot-terrain interaction forces for a trajectory pose,
it is possible to estimate the cost of each trajectory.
We decide to calculate the cost as a values variance of reaction forces acting on the robot along a trajectory $\tau$:
\begin{equation}
    \label{eq:traj_cost}
    \mathcal{C}_{\tau} = \frac{1}{T}\sum_{t=0}^{T}\left\|\mathbf{N}_t - \mathbf{\bar{N}}\right\|,
\end{equation}
where $\mathbf{N}_t$ is the predicted reaction force acting on the robot at a time moment $t$
and $\mathbf{\bar{N}}$ is the mean value of the total reaction forces along the trajectory.
Note that in the~\eqref{eq:traj_cost}, we denote
by $\mathbf{N}_t$ the total reaction force acting on all robot's body points at a time moment $t$.
The waypoint cost for a trajectory $\tau$ is simply calculated as the Euclidean distance between the trajectory and the next waypoint:
\begin{equation}
    \label{eq:waypoint_cost}
    \mathcal{C}_{\text{wp,$\tau$}} = \min_{x \in \tau}\left\|\mathbf{x} - \mathbf{x}_{\text{wp}}\right\|,
\end{equation}

The total cost used for the trajectory selection in navigation is the weighted sum of the trajectory and waypoint costs:
\begin{equation}
    \label{eq:total_cost}
    \mathcal{C}_{\text{total}} = \alpha \mathcal{C}_{\tau} + \beta \mathcal{C}_{\text{wp,$\tau$}},
\end{equation}
where $\alpha$ and $\beta$ are the hyperparameters for the trajectory and waypoint costs, respectively.
With the described navigation approach, the robot is able to autonomously navigate in the forest environment with uneven terrain,
avoiding obstacles and following the set of waypoints, Fig.~\ref{fig:nav_monoforce}.
The Fig.~\ref{fig:nav_maps} shows the constructed point cloud map of the forest environment during the navigation experiment.
It also contains an aerial photo of the forest area with the robot's trajectory and the waypoints.
The robot was able to autonomously navigate in the forest environment and traverse the 260-meter-long path.

\section{Conclusion}
In this work, we have presented \emph{FusionForce}, an explainable, physics-informed,
and end-to-end differentiable model that predicts the robot's trajectory from monocular camera images and/or lidar pointclouds.
The proposed \emph{grey-box} model combines classical learnable layers with the neural-symbolic physics-aware layer, which allows for significantly better generalization, especially in out-of-distribution terrains, than purely data-driven baselines. Consequently, it was demonstrated that our physics-based approach
($\nabla$Physics) significantly outperforms its data-driven baseline (TrajLSTM~\cite{pang2019aircraft})
and achieves $2$-times smaller error in trajectory prediction. We emphasize that the inherent absence of hazardous trajectories in training data remains an unresolved issue that will need to be addressed in future by adding some synthetic (visually hyper-realistic) data from modern game engines.

The model also provides interpretable intermediate outputs, such as friction or contacts, which can serve as an additional source of supervision. If such quantities are not directly measurable, it may still be possible to construct PINN-like losses to provide some further regularization. 

The training process is self-supervised; it only requires monocular camera images,
lidar scans, and SLAM-reconstructed trajectories.
In order to speed up the training process, we also leverage the foundation model for semantic segmentation that delivers rigid classes such as stones or tree trunks.

% MAY BE SOMWHERE HERE OR DELETE COMPLETELY:
% In our implementation,
% the black-box model predicts robot-terrain interaction forces and the true height at which they appear,
% %shape of the robot-supporting terrain,
% while the neural symbolic layer, which contains a differentiable physics engine, solves the robot's trajectory
% by querying these forces at the robot-terrain contacts.
% The main advantage of this approach is that it enables further self-supervision induced by lidar measurements,
% which essentially provides an
% % differentiable
% upper bound on the terrain shape.
% Since lidar measurements are not restricted to safe terrains,
% the prediction of the shape remains statistically consistent.
% If we could further assume that terrain textures on unsafe terrains also appear on safe terrains,
% the whole procedure would be statistically consistent.
% However, this is not true in general, as there could be some inherently unsafe textures,
% such as holes covered by vegetation,
% which do not have their safe counterpart.

\section*{Acknowledgments}
This work was co-funded by the Grant Agency of the CTU in Prague under Project SGS24/096/OHK3/2T/13, the European Union under the project Robotics and advanced industrial production (Roboprox reg. no. CZ.02.01.01/00/22\_008/0004590), the Czech Science Foundation under Project 24-12360S.

\bibliographystyle{IEEEtran}
\bibliography{IEEEabrv, references}

\end{document}